\documentclass[lettersize,journal]{IEEEtran}
\usepackage{amsmath,amsfonts}
\usepackage{algorithmic}
\usepackage{algorithm}
\usepackage{array}
\usepackage{textcomp}
\usepackage{stfloats}
\usepackage{url}
\usepackage{verbatim}
\usepackage{graphicx}
\usepackage{cite}
\usepackage{multirow}
\usepackage{booktabs}
\usepackage{float}
\usepackage{xcolor}
\usepackage{amsmath}
\usepackage{makecell}
\usepackage{array}

\usepackage{subcaption}
\RequirePackage{subcaption}

\usepackage{ragged2e}
\usepackage{booktabs}  
\usepackage{graphicx}  
\newcolumntype{Y}{>{\RaggedRight\arraybackslash}X}

\definecolor{darkgreen}{rgb}{0.0, 0.5, 0.0}

\usepackage{pifont}   

\usepackage{hyperref}

\begin{document}

\title{LibContinual: A Comprehensive Library towards Realistic Continual Learning}

\author{Wenbin Li, Shangge Liu, Borui Kang, Yiyang Chen, KaXuan Lew, Yang Chen, \\ Yinghuan Shi, Lei Wang, \textit{Senior Member, IEEE}, Yang Gao and Jiebo Luo, \textit{Fellow, IEEE}
\thanks{
Wenbin Li, Shangge Liu, Borui Kang, Yiyang Chen, KaXuan Lew, Yang Chen, Yinghuan Shi and Yang Gao are with the State Key Laboratory for Novel Software Technology, Nanjing University, China (e-mails: liwenbin@nju.edu.cn; lshangge@smail.nju.edu.cn; kangborui@smail.nju.edu.cn; yychen@smail.nju.edu.cn; lewkaxuan@gmail.com; chen-yang@smail.nju.edu.cn; syh@nju.edu.cn; gaoy@nju.edu.cn).}
\thanks{
Lei Wang is with the School of Computing and Information Technology, University of Wollongong, Australia (e-mail: leiw@uow.edu.au).
}
\thanks{
Jiebo Luo is with the Department of Computer Science, University of Rochester, USA (e-mail: jluo@cs.rochester.edu).
}
}

\markboth{}%
{Shell \MakeLowercase{\textit{et al.}}: A Sample Article Using IEEEtran.cls for IEEE Journals}


\maketitle

\begin{abstract}
A fundamental challenge in Continual Learning (CL) is catastrophic forgetting, where adapting to new tasks degrades the performance on previous ones. While the field has evolved with diverse methods, this rapid surge in diverse methodologies has culminated in a fragmented research landscape. The lack of a unified framework, including inconsistent implementations, conflicting dependencies, and varying evaluation protocols, makes fair comparison and reproducible research increasingly difficult. To address this challenge, we propose LibContinual, a comprehensive and reproducible library designed to serve as a foundational platform for realistic CL. Built upon a high-cohesion, low-coupling modular architecture, LibContinual integrates 19 representative algorithms across five major methodological categories, providing a standardized execution environment. Meanwhile, leveraging this unified framework, we systematically identify and investigate three implicit assumptions prevalent in mainstream evaluation: (1) offline data accessibility, (2) unregulated memory resources, and (3) intra-task semantic homogeneity. We argue that these assumptions often overestimate the real-world applicability of CL methods. Through our comprehensive analysis using strict online CL settings, a novel unified memory budget protocol, and a proposed category-randomized setting, we reveal significant performance drops in many representative CL methods when subjected to these real-world constraints. 
Our study underscores the necessity of resource-aware and semantically robust CL strategies, and offers LibContinual as a foundational toolkit for future research in realistic continual learning. The source code is available from \href{https://github.com/RL-VIG/LibContinual}{https://github.com/RL-VIG/LibContinual}.
\end{abstract}


\section{Introduction}\label{sec:intro}
\IEEEPARstart{E}{ndowing} machines with the human-like capability to continuously acquire new knowledge and adapt to evolving environments represents one of the ultimate milestones toward achieving artificial general intelligence. Continual Learning (CL), also known as lifelong or incremental learning, is the research paradigm dedicated to this goal. It requires a model to learn sequentially from a non-stationary stream of data, acquiring knowledge from a series of tasks without compromising its performance on previously learned tasks. However, this ideal learning process is hindered by a fundamental challenge, catastrophic forgetting~\cite{McCloskey1989},\cite{FRENCH1999}. When a model adjusts its parameters to accommodate a new task's data distribution, it often overwrites the knowledge critical for past tasks, causing significant performance degradation on prior tasks. Navigating the trade-off between plasticity (the ability to learn new knowledge) and stability (the preservation of old knowledge), known as the stability-plasticity dilemma~\cite{mermillod2013}, constitutes the central challenge in the field.

To address this challenge, the research community has explored various technical avenues~\cite{Wang2024}, including regularization-based methods~\cite{ewc2017},\cite{lwf2016} that protect prior knowledge, replay-based methods~\cite{icarl2017},\cite{der2021},\cite{van2020} that rehearse past data, optimization-based methods~\cite{gpm2021},\cite{inflora2024} that constrain parameter updates, and architecture-based methods~\cite{sdlora2025},\cite{moe4cl2024} that adapt the model structure. More recently, the advent of Pre-Trained Models (PTM) has triggered a profound paradigm shift in CL, giving rise to representation-based methods~\cite{l2p2022},\cite{ranpac2023},\cite{rapf2024} that focus on efficiently adapting powerful pre-trained features. The research focus is gradually moving from ``learning from scratch"~\cite{icarl2017} to ``efficiently and robustly fine-tuning and adapting powerful pre-trained knowledge"~\cite{l2p2022},\cite{Wang2024}. 

Meanwhile, this rapid surge in diverse methodologies has culminated in a fragmented research landscape.
Methods are often implemented using different deep learning frameworks, conflicting dependency versions, and inconsistent data processing pipelines. Such fragmentation makes it difficult to determine whether performance gains stem from algorithmic innovation or merely from differences in implementation details and hyperparameter tuning.
\begin{figure*}[t]
  \centering
   \includegraphics[width=\linewidth]{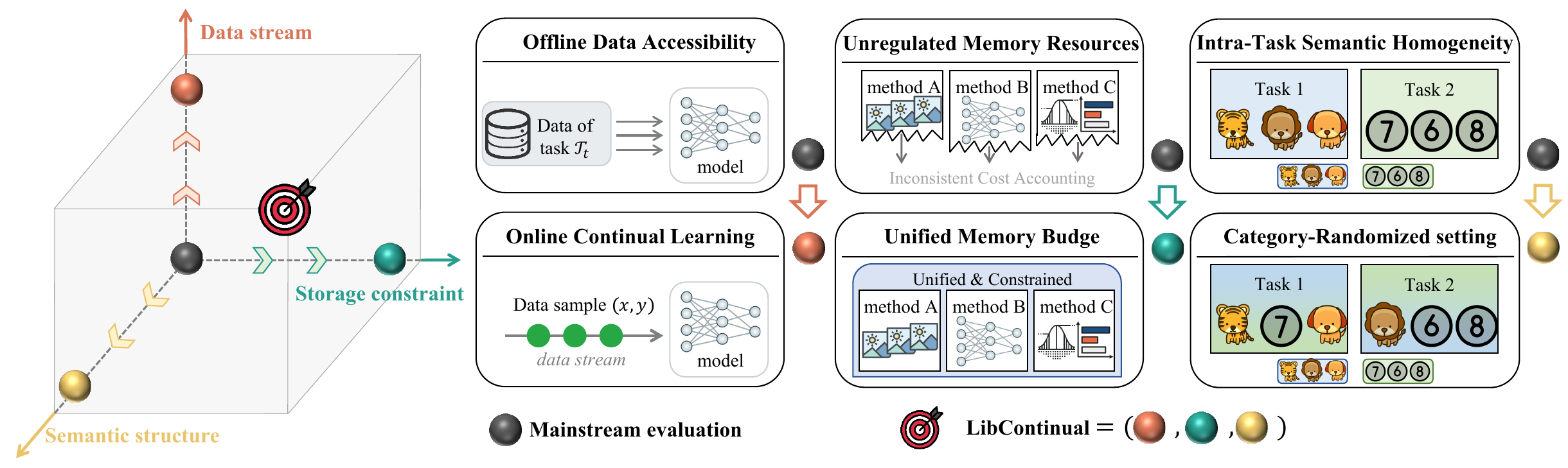}
   \caption{Conceptual illustration of the three implicit assumptions in continual learning evaluation and our proposed experimental dimensions to investigate them. Standard evaluation paradigms (gray) often rely on idealized conditions. (1) Data Stream (the Orange Axis): They assume offline access to task data for multi-epoch training, whereas we test models in a strict single-pass online CL setting. (2) Storage Constraint (the Teal Axis): They permit inconsistent memory cost accounting, making fair comparison difficult; we enforce a unified memory budget to normalize evaluation. (3) Semantic Structure (the Yellow Axis): They typically use tasks with high intra-task semantic homogeneity. We introduce a more challenging category-randomized setting, preventing models from relying on task-level shortcuts and thus testing for more robust representations.}
   \label{fig:challenge-setting}
\vspace{-10pt}
\end{figure*}
Consequently, there is a critical lack of a unified framework capable of providing a standardized implementation and fair comparison across the diverse methods. While several libraries have been developed to aid reproducible research, such as Avalanche~\cite{avalanche}, Continuum~\cite{continuum} and PyCIL~\cite{pycil}, they often exhibit limitations. 
As will be discussed in Section~\ref{subsec:relatedwork}, some existing frameworks suffer from rigid component coupling, which complicates the extension or customization of internal modules.
Others lack native support for modern Vision-Language Models (VLM), restricting the comparison between traditional training-from-scratch methods and PTM-based strategies. This absence of a comprehensive, modular, and up-to-date toolkit creates a barrier to rigorous empirical analysis and further advances. 

To address this challenge, we propose LibContinual, a comprehensive and reproducible library designed to serve as a foundational platform for realistic continual learning. LibContinual is built upon a high-cohesion, low-coupling architectural design (detailed in Section~\ref{sec:framework}). It decouples the experimental workflow into modular components, inlcuding Trainer, Model, Buffer, and DataModule, driven by a unified configuration system. 
This design allows researchers to seamlessly mix and match diverse backbones, classifiers, and buffer strategies within a standardized execution environment. 
Leveraging this architecture, we integrate 19 representative algorithms spanning all five major categories: regularization, replay, optimization, representation, and architecture-based methods. By providing a unified interface for classical and modern PTM-based methods, LibContinual can enable the community to conduct fair, transparent, and scalable benchmarking.

Equally important, during the development of LibContinual, the standardization of protocols allowed us to identify several implicit assumptions deeply embedded in mainstream evaluation paradigms. These assumptions, often accepted as convention, may overestimate the real-world applicability of CL methods. Specifically, we identify three implicit assumption: (1) The Assumption of Offline Data Accessibility, which presumes multi-epoch training on task data, ignoring the single-pass nature of real-world streams; (2) The Assumption of Unregulated Memory Resources, where inconsistent accounting of storage (\textit{e.g.}, raw images vs. abstract features) obscures true algorithmic efficiency; and (3) The Assumption of Intra-Task Semantic Homogeneity, which provides models with artificial contextual shortcuts by grouping semantically related classes into tasks.

Leveraging the modular capabilities of LibContinual, we move beyond simple benchmarking to systematically investigate these assumptions. As conceptually illustrated in Figure~\ref{fig:challenge-setting}, our investigation is structured along the three dimensions. We introduce novel evaluation protocols, including a strict online CL setting, a unified memory budget, and a challenging category-randomized setting. 
Our comprehensive experiments reveal that when these idealized assumptions are removed, the performance of some methods degrades significantly, exposing the fragility of current solutions under realistic constraints.

The contributions of this paper are summarized as follows:
\begin{itemize}
\item[$\bullet$] 
We propose the LibContinual, a unified and reproducible framework designed for the rigorous and fair implementation and evaluation of continual learning algorithms.

\item[$\bullet$] 
We systematically identify and investigate three fundamental yet often overlooked assumptions in mainstream CL evaluation: offline data accessibility, unregulated memory resources, and intra-task semantic homogeneity.

\item[$\bullet$] We propose novel evaluation protocols, including a unified memory budget and a category-randomized setting, to facilitate more realistic and robust benchmarking.

\item[$\bullet$] Through extensive experiments within LibContinual, we reveal significant performance drops for many representative methods under more realistic settings. 
Our findings provide critical insights into their true applicability and underscore the necessity of developing resource-aware and semantically robust CL strategies.
\end{itemize}
\section{Background and Related Work}
\subsection{The Continual Learning Problem}\label{subsec:problem formulation}
Continual Learning is the paradigm for training models on a sequence of tasks, where the data distribution is non-stationary. 
The central challenge is to overcome catastrophic forgetting, where a model’s performance on previously learned tasks degrades significantly upon learning new ones.

Formally, the problem of continual learning is defined as a process of sequential learning over a sequence of tasks, $\boldsymbol{\mathcal{T}} = \{\mathcal{T}_1, \mathcal{T}_2, \dots, \mathcal{T}_T\}$. Each task $\mathcal{T}_t$ is characterized by a distinct data distribution $\mathcal{D}_t$ over an input space $\mathcal{X}_t$ and a label space $\mathcal{Y}_t$. A learning system, represented by a model $f_{\theta}$ parameterized by $\theta$, learns from this sequence of tasks in order. When learning the $t$-th task, the model updates its parameters from $\theta_{t-1}$ to $\theta_t$ based on data sampled from $\mathcal{D}_t$.

The core constraint of continual learning is that while learning task $\mathcal{T}_t$, the model has very limited access to the training data from past tasks' distributions $\{\mathcal{D}_1, \dots, \mathcal{D}_{t-1}\}$. The ultimate objective is to find a single set of parameters $\theta_T$ for the final model that minimizes the total statistical risk across all tasks seen so far,
\begin{equation}\small
    \min_{\theta_T}\sum_{t=1}^T \mathbb{E}_{(x,y)\sim\mathcal{D}_t} [\mathcal{L}(f_{\theta_T}(x),y)],
\end{equation}
where $\mathcal{L}$ is a given loss function. However, this objective is infeasible in continual learning because restricted access to past data renders the global empirical risk incomputable. This inability to evaluate the empirical risk on previous tasks is the root cause of catastrophic forgetting.

\subsection{Related Work}\label{subsec:relatedwork}

While several excellent works have mapped the continual learning landscape, our work offers a distinct and targeted contribution. We position LibContinual by comparing it first against broad academic surveys and then against existing software libraries.

\subsubsection{Comparison with Continual Learning Surveys}

A significant body of literature provides comprehensive surveys of the continual learning field. 
Lange et al.~\cite{Lange2022}, Wang et al.~\cite{Wang2024}, and Zhou et al.~\cite{Zhou2024} offer extensive taxonomies. They categorize methods into families such as regularization-based, replay-based, and architecture-based methods. These surveys are invaluable for understanding theoretical underpinnings and the historical progression of algorithms. More focused surveys also exist for specific sub-fields. For example, Masana et al.~\cite{Masana2023} focus on class-incremental learning, while Zhou et al.~\cite{zhou24pretrain} provide deep dives into pre-trained models. These works successfully synthesize and organize existing knowledge.

However, these surveys are primarily descriptive and rely on results from inconsistent experimental paradigms. 
These paradigms often rest on idealized premises that may not hold in realistic scenarios. 
In contrast, our contribution shifts from a descriptive role to an \textit{experimental and prescriptive} one. 
We implement concrete protocols to scrutinize three prevalent implicit assumptions: \textit{offline data accessibility}, \textit{intra-task semantic homogeneity}, and \textit{unregulated memory resources}. 
By rigorously investigating these factors, we complement theoretical surveys with robust practical verification.

\subsubsection{Comparison with Continual Learning Libraries}

Several open-source libraries have been developed to aid reproducible research in continual learning. Avalanche~\cite{avalanche} is a comprehensive library offering a vast collection of algorithms and standard benchmarks. Continuum~\cite{continuum} excels with its robust and flexible data-loading capabilities. Other libraries like PyCIL~\cite{pycil} provide a focused toolbox specifically for class-incremental learning. These frameworks have been instrumental in standardizing experiments. 

While these frameworks help standardize experiments, LibContinual advances distinct advancements in architectural design and model compatibility.
First, LibContinual features a \textit{unified and decoupled high-level design}. We adhere to high-cohesion and low-coupling principles by extracting algorithm-agnostic components into shared modules. 
These components include the training loop, data management, and evaluation protocols. 
Consequently, the implementation files remain minimal. 
Researchers only need to define functions specific to the unique logic of an algorithm. This modularity reduces code redundancy and guarantees a standardized execution environment for fair comparison.
Second, LibContinual explicitly supports \textit{Vision-Language Models (VLM)}. 
The framework seamlessly integrates modern backbones such as Vision Transformers and CLIP. 
This capability allows for the direct implementation and comprehensive evaluation of advanced PTM-based strategies alongside classical methods.

\section{LibContinual: A Unified Framework for Continual Learning}\label{sec:framework}

\begin{figure}[t]
  \centering
   \includegraphics[width=\linewidth]{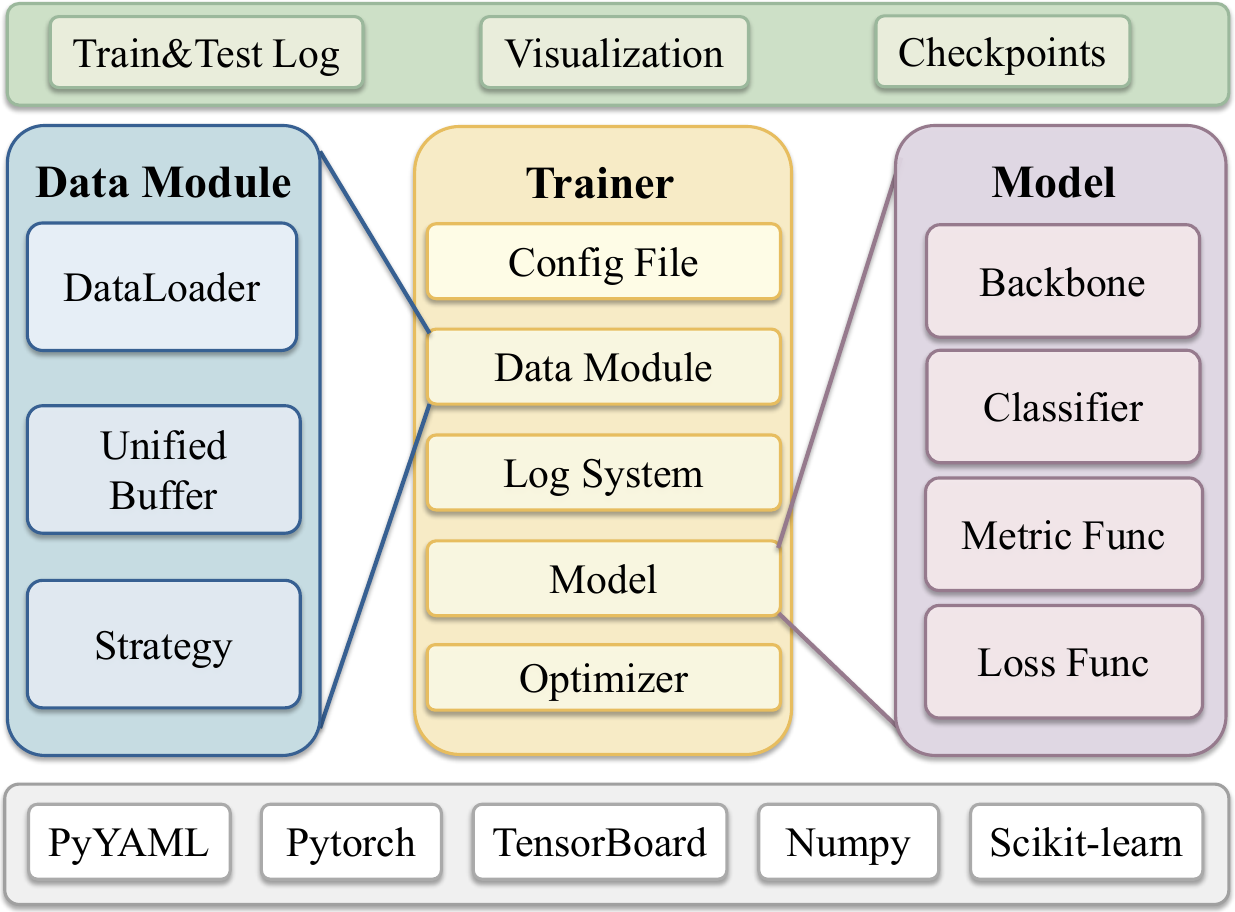}
   \caption{
   Architecture of the proposed Libcontinual. 
   }
   \label{fig:libcl}
\vspace{-10pt}
\end{figure}

\subsection{The LibContinual Toolbox}
The CL landscape is characterized by a rapid expansion of algorithms. However, this progress is often fragmented across disparate codebases with inconsistent evaluation protocols, making fair comparison and reproducible research significant challenges. To address this critical gap, we propose LibContinual. It is not merely an algorithm repository but a comprehensive and modular research toolbox built upon PyTorch. It is designed to provide a standardized environment that fosters transparent, rigorous, and fair algorithmic evaluation.

As shown in Figure~\ref{fig:libcl}, LibContinual adheres to high-cohesion, low-coupling design principles and is driven by simple YAML files for experiment configuration. Its architecture is meticulously decoupled into several core modules: a Trainer module that manages the entire experimental workflow; a Model module integrating diverse backbones, classifiers, and algorithms; a specialized DataLoader for CL-specific data partitioning and augmentation; a versatile Buffer module supporting various memory sampling and update strategies; and a Config module that orchestrates the entire setup.

This clear modularity and unified interface are designed to provide researchers with a fair evaluation platform while significantly lowering the barrier to developing and validating novel algorithms. We believe LibContinual can serve as a robust foundation for advancing the field. A detailed overview of the framework's architecture, module functionalities, and implementation specifics is shown in Appendix B.

\subsection{Continual Learning Scenarios Supported in Libcontinual}\label{subsec:classic_classification}


Continual learning scenarios are typically categorized along several distinct, often orthogonal, dimensions. These dimensions determine the specific constraints and challenges the learning algorithm needs to address. 
A key design goal of LibContinual is to provide a unified platform that supports the configuration of these diverse scenarios via modular components and YAML files, enabling systematic performance analysis. In the following, we explore three of these fundamental dimensions, all configurable within our framework: the data arrival paradigm, the information accessible at inference time, and the semantic structure of the tasks.

\subsubsection{By data arrival paradigm} \label{subsubsec:online_batch}
This dimension defines how the data for an individual task $\mathcal{T}_t$ is presented to the model. LibContinual supports the two dominant paradigms.

In \textbf{Offline Continual Learning} (or batch CL), the entire training dataset for a task, $D_t=\{(x_i, y_i)\}_{i=1}^{n_t}$, is made available at once. The model parameters are updated from $\theta_{t-1}$ to $\theta_t$ by optimizing over the full training dataset $D_t$\footnote{Here, $D_t$ represents the finite, empirical dataset sampled from the underlying theoretical data distribution $\mathcal{D}_t$ introduced in Section~\ref{subsec:problem formulation}. While the ultimate goal is to generalize to $\mathcal{D}_t$, the learning algorithm only has access to $D_t$.}, often for multiple epochs. Repeated training ensures the model thoroughly converges on the objectives for task $\mathcal{T}_t$ before moving on. The learning process for the task can be abstracted as $\theta_t =\mathrm{Train}(\theta_{t-1},{D}_t)$.

In \textbf{Online Continual Learning}, data arrives as a continuous and often rapid stream, demanding that the model learns on-the-fly. The data is typically processed as a sequence of small mini-batches $(B_{t,1}, B_{t,2}, \dots, B_{t,K_t})$, where $D_t = \bigcup_{k=1}^{K_t} B_{t,k}$. The model is restricted to a single pass, meaning parameters are updated incrementally after each mini-batch,
\begin{equation}\small
\theta_{t,k} = \mathrm{Update}(\theta_{t,k-1}, B_{t,k}), \quad \text{for } k=1, \dots, K_t \label{eq:online_update} 
\vspace{-4pt}
\end{equation}
where $\theta_{t,0} = \theta_{t-1}$ and the final parameter set is $\theta_t = \theta_{t,K_t}$. This single-pass constraint gives rise to intra-task forgetting~\cite{ocm2022}, the tendency to forget knowledge from earlier batches while learning from later ones.

This distinction is configured via the YAML file. Consistent with online CL protocols~\cite{ocm2022,erace2022}, users can enforce a strict single-pass stream by setting `\textit{epochs: 1}' and specifying a small batch size (\textit{e.g.}, 10). Conversely, setting `\textit{epochs $>$ 1}' enables the multi-epoch training.

\subsubsection{By inference-time accessible information}\label{subsubsec:til_dil_cil}
The second dimension classifies scenarios based on the information available at test time, leading to the three canonical settings proposed by~\cite{vande2022}: Task-Incremental, Domain-Incremental, and Class-Incremental Learning.

In \textbf{Task-Incremental Learning}, the model is provided with the task identity $t$ at inference time. The objective is thus to learn a task-aware mapping $f_{\theta}: (\mathcal{X}, t) \to \mathcal{Y}_t$. This allows for task-specific components (\textit{e.g.}, a multi-headed classifier), shifting the primary challenge from merely preventing forgetting to achieving efficient knowledge transfer.

In contrast, both Domain- and Class-Incremental Learning operate without task identity at test time. In \textbf{Domain-Incremental Learning} (DIL), all tasks share an identical label space, \textit{i.e.}, $\mathcal{Y}_1 = \dots = \mathcal{Y}_T = \mathcal{Y}_{\text{shared}}$. The model must learn a single, unified mapping $f_{\theta}: \mathcal{X} \to \mathcal{Y}_{\text{shared}}$ that is robust to shifts in the input data distribution (domains).

Finally, in \textbf{Class-Incremental Learning}, each task introduces a new, disjoint set of classes, where $\mathcal{Y}_t \cap \mathcal{Y}_{t'} = \emptyset$ for any $t \neq t'$. The model must learn to discriminate among all classes seen so far, requiring a mapping to the global label space, $f_{\theta}: \mathcal{X} \to \bigcup_{i=1}^T \mathcal{Y}_i$. This is widely considered the most challenging scenario as it requires distinguishing between classes that are never observed together~\cite{vande2022}. 
Since guaranteeing the availability of task identity is often impractical in real-world environments, and CIL is considered a more challenging scenario~\cite{Wang2024}, we focus our investigation primarily on this setting.

\begin{figure}[t]
  \centering
   \includegraphics[width=\linewidth]{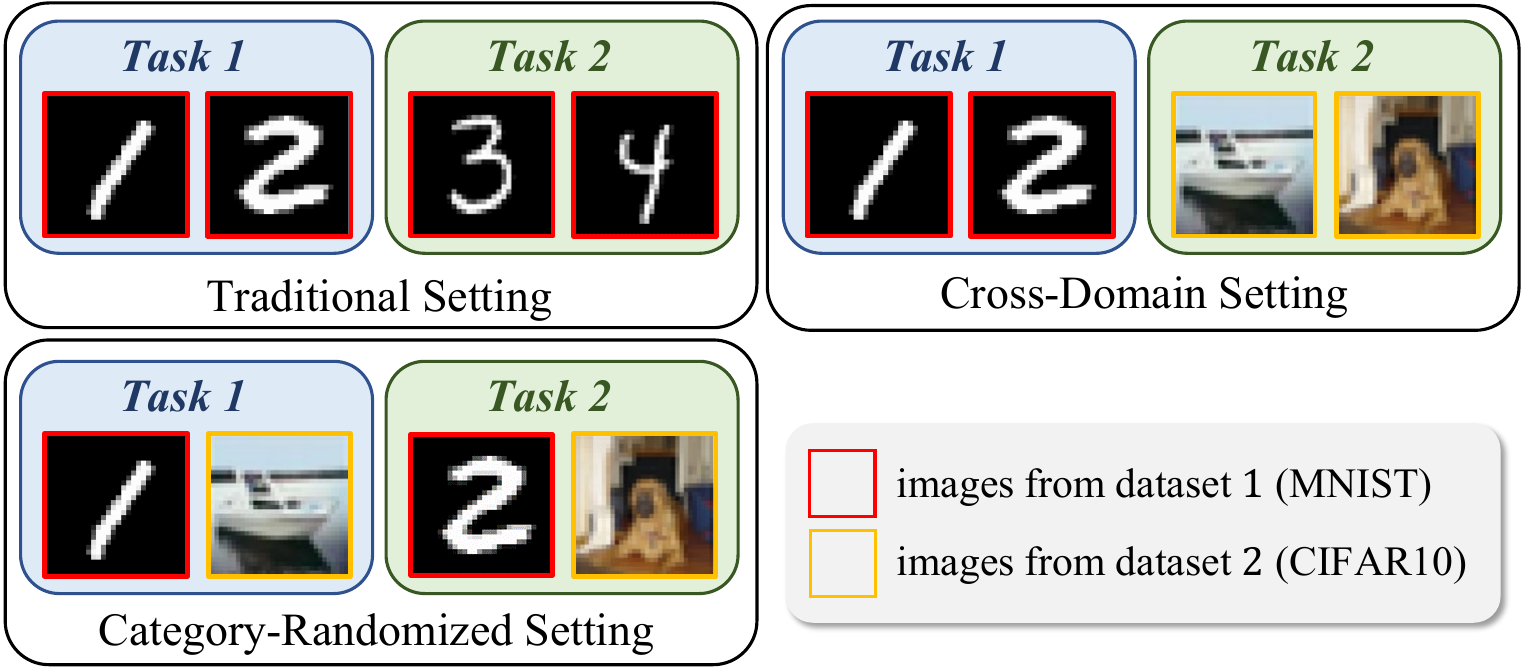}
   \caption{An illustration of the three continual learning settings defined by inter- and intra-task semantic structure. In the traditional setting, tasks are subsets of a single dataset, making them semantically homogeneous. In the cross-domain setting, each task remains homogeneous but originates from a different domain, introducing a domain shift between tasks. In our proposed category-randomized setting, the assumption of intra-task homogeneity is broken. Each task is a heterogeneous mix of classes from different domains, forcing the model to learn disparate concepts simultaneously.}
   \label{fig:setting_datacomposition}
\vspace{-10pt}
\end{figure}

LibContinual provides robust support for the two most widely studied scenarios: TIL and CIL. In the YAML file, setting `\textit{setting: task-aware}' enables the TIL scenario, and setting `\textit{setting: task-agnostic}' enforces the CIL scenario.

\begin{figure*}[t]
  \centering
   \includegraphics[width=\linewidth]{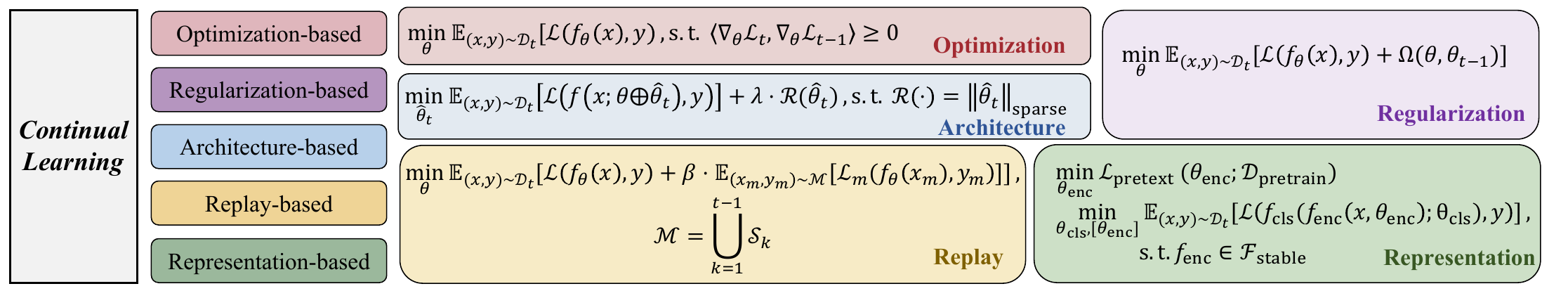}
   \caption{
   The taxonomy of continual learning methods categorizes them into five major algorithmic strategies: regularization-based, replay-based, optimization-based, representation-based, and architecture-based methods.
   }
   \label{fig:class_classification}
\end{figure*}

\subsubsection{By inter- and intra-task semantic structure}\label{subsubsec:Semantic Structure}

A third crucial dimension, which we introduce in this work to systematically analyze semantic assumptions, is the Inter- and Intra-Task Semantic Structure. It defines how the disjoint class sets ($\mathcal{Y}_t$) are constructed to form the sequence of tasks in a CIL problem. This construction governs two key properties of the learning curriculum: the intra-task semantic similarity (whether classes within a single task are related) and the inter-task semantic similarity (how related the consecutive tasks are). LibContinual supports three key setting along this dimension as illustrated in Figure~\ref{fig:setting_datacomposition}.

\textbf{Traditional Setting.} This setting is characterized by high intra-task and high inter-task semantic similarity. Tasks are typically formed by partitioning the classes of a single and semantically coherent dataset (\textit{e.g.}, splitting CIFAR-100 into 10 tasks). Consequently, classes within any given task are inherently related, and all tasks are drawn from the same overarching data distribution. This common setup allows models to leverage task-level semantic context but fails to represent learning across disparate concepts.

\textbf{Cross-domain Setting.} This setting features high intra-task but low inter-task semantic similarity. Each task remains internally homogeneous as all its classes are sourced from a single dataset, but different tasks originate from distinct domains, introducing a significant domain shift. For instance, a learning sequence might start with Task 1 containing classes from CIFAR-10~\cite{cifar} (natural images), followed by Task 2 with classes from MNIST~\cite{mnist} (handwritten digits). Although framed as a CIL problem with disjoint classes, the primary challenge here is the abrupt change in data statistics and visual features between tasks. This setup rigorously tests a model's ability to adapt to new data types while preserving knowledge from entirely different past domains.

\textbf{Category-randomized Setting.} 
LibContinual supports a novel category-randomized setting. Unlike the previous two paradigms which maintain high intra-task semantic homogeneity, this setting is designed to deliberately disrupt this semantic structure to rigorously test model robustness. We reserve the detailed formulation, motivation, and implementation details of this setting for Section~\ref{subsec:assumption-semantic}, where we discuss it in the context of investigating implicit semantic assumptions.



In LibContinual, these scenarios are configured via the data module. 
For the traditional setting, users specify a single dataset and a partitioning scheme. 
For the cross-domain setting, users can define a sequence of datasets, where each dataset is treated as a new task. Finally, the category-randomized setting is enabled by a flag that pools, shuffles, and partitions classes from all specified datasets.

\subsection{The Selection of Algorithms for Libcontinual}
\label{subsection:implemented}
To ensure a comprehensive and representative evaluation of the continual learning landscape, we have implemented a suite of 19 methods within LibContinual (see Table~\ref{tab:method_classification}). Our selection is guided by recent and widely accepted taxonomies in the field~\cite{Wang2024}, which categorize CL algorithms into five major families: regularization-based, replay-based, optimization-based, representation-based, and architecture-based strategies. By integrating both classical baselines and representative PTM methods, LibContinual facilitates a rigorous investigation into how different algorithmic philosophies address the stability-plasticity dilemma.

Crucially, to provide a cohesive theoretical perspective, we distill the core logic of each category into a mathematical formulation as shown in Figure~\ref{fig:class_classification}. For each category, we will: (1) present its core principles through our distilled formulation, (2) highlight representative methods implemented in LibContinual, and (3) provide critical discussion on its distinctive advantages and unresolved limitations. Detailed descriptions, implementation specifics, and additional representative methods for each category are provided in Appendix A.

\subsubsection{Regularization-based methods}
These methods augment the training objective with a penalty term to mitigate forgetting. When learning task $\mathcal{T}_t$, the parameters $\theta_t$ are found by optimizing,
\begin{equation}\small
    \theta_t = \arg\min_{\theta} \mathbb{E}_{(x,y)\sim\mathcal{D}_t}[\mathcal{L}(f_{\theta}(x),y) + \Omega(\theta,\theta_{t-1})].
\end{equation}
The objective balances plasticity, driven by the empirical risk on the new data $\mathcal{D}_t$, with stability, enforced by the regularizer $\Omega(\theta, \theta_{t-1})$ which penalizes deviations from the previous state. 
The primary innovation in this category lies in the specific design of the regularizer $\Omega$.

In early methods, the regularizer $\Omega$ can be defined in the parameter space~\cite{ewc2017},\cite{MAS2017},\cite{SI2017},\cite{RWALK2018} or functional space~\cite{lwf2016},\cite{Iscen2020},\cite{Castro2018},\cite{Triki2017}. This principle continues to evolve, with recent work exploring regularization in the spectral~\cite{Lewandowski2025} and topological~\cite{Fan2024} domains. Furthermore, researchers have also begun to adapt these regularization methods to the continual instruction tuning of large multimodal models~\cite{he2023, Zheng2023zscl}.

Regularization-based methods are foundational to continual learning due to their simplicity and effectiveness. 
However, they face challenges in the era of Pre-trained Models (PTM). 
Directly applying global constraints often proves suboptimal as it restricts the PTM's inherent adaptability~\cite{he2023}.
We argue that the future value of regularization lies in its flexibility as a strategic component, for instance, applying targeted constraints to lightweight modules like Adapters, rather than as a standalone global solution.

Within this category, LibContinual implements foundational methods including LwF~\cite{lwf2016} and EWC~\cite{ewc2017}, which regularize the model in the functional and parameter space, respectively.

\subsubsection{Replay-based methods}
These methods mitigate catastrophic forgetting by rehearsing a subset of past data stored in a memory buffer $\mathcal{M}$ alongside new task data, thereby approximating training on the joint distribution. The optimization objective minimizes a composite loss,
\begin{equation}\small
\begin{split}
    \min_{\theta}\mathbb{E}_{(x,y)\sim\mathcal{D}_t}[\mathcal{L}(f_{\theta}(x),y)]&+\beta\cdot\mathbb{E}_{(x_m,y_m)\sim\mathcal{M}}[\mathcal{L}_{m}(f_{\theta}(x_m),y_m)], \\ 
    &\mathcal{M}=\bigcup_{k=1}^{t-1}\mathcal{S}_k. \label{formula:replay}
\end{split}
\end{equation}
Here, $\mathcal{M}$ contains exemplars $\mathcal{S}_k$ from past tasks. The content of this buffer $\mathcal{M}$, fundamentally defines the specific replay strategy. It may contain raw past examples~\cite{icarl2017}~\cite{Oord2017}\cite{aljundi2019}\cite{bang2021}\cite{liu2020}, pseudo-data synthesized by a generative model~\cite{shin2017}\cite{ostapenko2019}\cite{cong2020}\cite{van2020}, or abstract latent representations~\cite{hayes2020}\cite{zhu2022}.
And the model is regularized by minimizing $\mathcal{L}_m$ on samples drawn from this buffer. The hyperparameter $\beta$ controls the trade-off between learning the new task and preserving old knowledge. 
The central challenge lies in how to construct, manage, and utilize the memory buffer 
$\mathcal{M}$ to approximate the true data distribution of past tasks under strict memory constraints.

Replay-based methods remain at the forefront of continual learning research due to the direct, data-driven constraint against forgetting. While effective, their performance is intrinsically tied to the size and quality of the memory buffer, creating a trade-off between memory overhead, potential privacy risks,  and learning efficacy. We argue that future progress hinges on two key areas: developing more intelligent sampling strategies to maximize the utility of a limited memory budget, and establishing a stronger theoretical foundation to explain why and when replay is most effective. 

In LibContinual, we implement representative methods including iCaRL~\cite{icarl2017}, BiC~\cite{bic2019}, LUCIR~\cite{lucir2019}, WA~\cite{wa2020}, and the online-focused OCM~\cite{ocm2022} and ERAML/ERACE~\cite{erace2022}.

\subsubsection{Optimization-based methods}
These methods mitigate the stability-plasticity dilemma by framing continual learning as a constrained optimization problem. The core idea is to find parameter updates for a new task that do not increase the loss on previously learned tasks. This principle is captured by the general objective,
\begin{equation}\small
\begin{aligned}\small
& \theta_t = \arg\min_{\theta} \left[ \mathbb{E}_{(x,y)\sim\mathcal{D}_t}\mathcal{L}(f_{\theta}(x),y)  \right]. 
& \text{s.t.} \langle \nabla_\theta \mathcal{L}_{t}, \nabla_\theta \mathcal{L}_{t-1} \rangle \geq 0
\end{aligned}
\end{equation}
The objective balances plasticity, by minimizing the loss on new data $\mathcal{D}_t$, with stability, enforced by the constraint that prevents conflicting updates. 

Classic methods like GPM~\cite{gpm2021} and Adam-NSCL~\cite{comatrix2021} enforce a strict constraint by projecting the gradient for the new task into a subspace that is orthogonal to the feature representations of past tasks. While effective at preventing interference, these hard constraints can limit plasticity. To address this, subsequent methods have introduced more flexible constraints. For instance, TRGP~\cite{trgp2022} uses adaptive trust regions to allow for beneficial updates within boundaries, while more recent work like AdaBOP~\cite{adbop2025} derives a closed-form solution for the optimal projection, enabling a fine-grained, per-layer balance between stability and plasticity via tunable hyperparameters. 

Optimization-based methods are often adopted as effective plug-and-play modules for mitigating forgetting. However, this has led to a focus on applying existing constraint techniques rather than innovating on the underlying optimization principles themselves~\cite{prompt_nscl2024}.
Many current methods still rely on variations of gradient projection~\cite{kang2025dynamic}. 
We believe future progress lies in re-examining the foundational theory to develop optimization frameworks that intrinsically encode forgetting resistance.

In LibContinual, we include prominent optimization-based methods such as GPM~\cite{gpm2021} and TRGP~\cite{trgp2022}.

\subsubsection{Representation-based methods}
These methods shift the CL focus from preserving old knowledge to acquiring universal representations. The core philosophy involves a two-phase process where the model decomposes into a feature encoder $f_\text{cls}$ and a classifier $\theta_\text{cls}$.

\begin{table}[t]
\centering
 \caption{Classification overview of implemented methods in LibContinual. This table summarizes the representative algorithms supported by our framework, classifying them by their core algorithmic Strategy (Section~\ref{subsec:classic_classification}) and the form of Storage they use to mitigate forgetting (Section~\ref{subsec:assumption_memory}).
}
\label{tab:method_classification}
\begin{tabular}{lccc}
\toprule
\textbf{Method} & \textbf{Venue \& Year} & \textbf{Algorithm} & \textbf{Storage} \\ 
\midrule
LwF~\cite{lwf2016}            & ECCV 2016       & Regularization                     & Feature \\
EWC~\cite{ewc2017}            & PNAS 2017       & Regularization                     & Model \\
\midrule
iCaRL~\cite{icarl2017}          & CVPR 2017       & Replay                             & Image \\
BiC~\cite{bic2019}            & CVPR 2019       & Replay                             & Image \\
LUCIR~\cite{lucir2019}          & CVPR 2019       & Replay                             & Image \\
WA~\cite{wa2020}             & CVPR 2020       & Replay                             & Image \\
ERAML/ERACE~\cite{erace2022}         & ICLR 2022       & Replay                             & Image \\
OCM~\cite{ocm2022}            & ICML 2022       & Replay                             & Image \\
\midrule
GPM~\cite{gpm2021}            & ICLR 2021       & Optimization                       & Feature \\
TRGP~\cite{trgp2022}           & ICLR 2022       & Optimization                       & Feature \\
\midrule
API~\cite{api2023}            & CVPR 2023       & Architecture                       & Feature \\
InfLoRA~\cite{inflora2024}        & CVPR 2024       & Architecture                       & Feature \\
MoE-Adapter4CL~\cite{moe4cl2024} & CVPR 2024       & Architecture                       & Parameter \\
SD-LoRA~\cite{sdlora2025}        & ICLR 2025       & Architecture                       & Parameter \\
\midrule
L2P~\cite{l2p2022}            & CVPR 2022       & Representation                     & Prompt \\
DualPrompt~\cite{dualprompt2022}     & ECCV 2022       & Representation                     & Prompt \\
CodaPrompt~\cite{codaprompt2023}     & CVPR 2023       & Representation                     & Prompt \\
RanPAC~\cite{ranpac2023}         & NeurIPS 2023    & Representation                     & Parameter \\
RAPF~\cite{rapf2024}           & ECCV 2024       & Representation                     & Parameter \\
\bottomrule
\end{tabular}%
\vspace{-15pt}
\end{table}

\noindent \textbf{Phase 1} (Representation Learning).
\begin{equation}\small
    \min_{\theta_\text{enc}} \mathcal{L}_{\text{pretext}}(\theta_\text{enc}; \mathcal{D}_{\text{pretrain}}).
\end{equation}

\noindent \textbf{Phase 2} (Continual Learning).
\begin{equation}\small
    \begin{split}
        \min_{\theta_\text{cls}, [\theta_\text{enc}]} & \mathbb{E}_{(x,y) \sim D_t} [\mathcal{L}(f_\text{cls}(f_\text{enc}(x; \theta_\text{enc}); \theta_\text{cls}), y)] \\
        & \text{s.t.} \quad f_\text{enc} \in \mathcal{F}_\text{stable}.
    \end{split}
\end{equation}
In the first phase, a powerful encoder $f_\text{enc}$ is trained on $\mathcal{D}_\text{pretrain}$ via a pretext task, optimizing a pretext loss $\mathcal{L}_\text{pretext}$. This phase, which typically involves self-supervised learning or large-scale supervised pre-training~\cite{clip2021,vit2021}. Most CL methods concentrate on the strategies for Phase 2, which focuses on adapting to task $\mathcal{T}_t$ while maintaining stability. Strategies to enforce $f_{\text{enc}} \in \mathcal{F}_{\text{stable}}$ generally fall into two categories: keeping $\theta_{\text{enc}}$ frozen while optimizing lightweight modules, such as prompts (L2P~\cite{l2p2022}, DualPrompt~\cite{dualprompt2022}, CodaPrompt~\cite{codaprompt2023}) or random projection layers (RanPAC~\cite{ranpac2023}), or cautiously fine-tuning the encoder to balance plasticity and forgetting~\cite{Zhang2023}\cite{Cha2021}\cite{rapf2024}.

Representation-based methods, especially those built upon PTM, currently represent one of promising frontiers of rehearsal-free continual learning. Their success highlights a key insight: the core challenge of CL is not learning new features from scratch, but rather learning to effectively access and combine the rich features already present in PTM. Therefore, the central challenges lie in learning more potent feature representations and developing more fine-grained mechanisms to identify and preserve the features essential for preventing forgetting. Furthermore, addressing the representation gaps and discrepancies between different modalities, and how to continually learn on them without catastrophic interference, emerges as a vital new frontier.

In LibContinual, we implement L2P~\cite{l2p2022}, DualPrompt~\cite{dualprompt2022}, CODA-Prompt~\cite{codaprompt2023}, RanPAC~\cite{ranpac2023}, and RAPF~\cite{rapf2024}.

\subsubsection{Architecture-based methods}
These methods prevent catastrophic forgetting by structurally isolating task-specific knowledge. They typically compose a stable, shared backbone $\theta$ with expandable, task-specific modules $\{\hat{\theta}_k\}_{k=1}^t$. The optimization for task $\mathcal{T}_t$ focuses on updating only the newly introduced parameters,
\begin{equation}\small
    \begin{split}
        \min_{\hat{\theta}_t} \mathbb{E}_{(x,y) \sim \mathcal{D}_t} [\mathcal{L}(f(x; & \theta \oplus \hat{\theta}_t), y)] + \lambda \cdot \mathcal{R}(\hat{\theta}_t)\\
        & \text{s.t.} \quad \mathcal{R}(\cdot)=||\hat{\theta}_t||_{\text{sparse}}.
    \end{split}
\end{equation}
Here, $\theta$ is typically frozen to ensure stability, while $\hat{\theta}_t$ provides plasticity. The operator $\oplus$ signifies how these parameter sets are combined, such as through masking, additive decomposition, or modular routing. $\mathcal{R}(\cdot)$ is often used to enforce constraints like sparsity, ensuring the model's growth is scalable.

Early methods focused on parameter isolation within a fixed-capacity model~\cite{Yoon2018}\cite{Serra2018}\cite{Mallya2018}\cite{Oswald2020}\cite{Aljundi2017}\cite{api2023}\cite{Serra2018}\cite{Mallya2018}. These methods, while effective, were often designed for models trained from scratch. With the advent of PTM, $\hat{\theta}_t$ is commonly realized through Parameter-Efficient Fine-Tuning (PEFT) techniques. 
Examples include constraining LoRA matrices~\cite{inflora2024,sdlora2025} or utilizing Mixture-of-Experts~\cite{moe4cl2024}.


Architecture-based methods address the stability-plasticity dilemma structurally by freezing the general knowledge base ($\theta$) while enabling targeted updates via $\hat{\theta}_t$. 
However, they face significant challenges in CIL setting, as selecting the correct task-specific parameters $\hat{\theta}_t$ without a task identity oracle is non-trivial. 
Moreover, as the number of tasks grows, naively accumulating task-specific parameters can lead to a linear or super-linear growth in model size. The future of this domain lies in developing more sophisticated strategies for managing $\hat{\theta}_t$. Instead of simple expansion, the focus is shifting towards intelligent parameter reuse, composition, and merging.

In LibContinual, we implement API~\cite{api2023}, InfLoRA~\cite{inflora2024}, MoE-Adapter4CL~\cite{moe4cl2024}, and SD-LoRA~\cite{sdlora2025}, representing both training-from-scratch and PTM-based methods.

\subsection{Components and Benchmarks Supported in Libcontinual}
\label{subsection:metrics}
To ensure comprehensive and reproducible experiments, LibContinual provides a modular suite of standardized components, backbones, and benchmarks.

\textbf{Backbone architectures.} The framework supports both classic Convolutional Neural Networks (CNNs) like AlexNet~\cite{alexnet}, ResNet-18, and ResNet-32~\cite{resnet2016}, as well as modern Pre-trained Models such as the Vision Transformer (ViT)~\cite{vit2021} and CLIP~\cite{clip2021}.

\textbf{Datasets and benchmarks.} We integrate widely-used datasets, including CIFAR-10, CIFAR-100, TinyImageNet, and the more challenging ImageNet-R. 
For evaluating domain adaptation, the framework also incorporates the standard 5-datasets cross-domain benchmark (comprising CIFAR-10, MNIST, Fashion-MNIST, SVHN, and notMNIST) specifically designed for evaluating domain adaptation capabilities in continual learning scenarios. 


\textbf{Evaluation metrics.} 
While various performance metrics have been proposed in continual learning to evaluate model performance, we adopt two standard metrics in our experiments due to their widespread adoption and the convenience they offer for comparative analysis~\cite{moe4cl2024}\cite{metric_1}.
Let $\mathcal{A}_{t,j} \in [0,1]$ denote the accuracy evaluated on the test set of the $j$-th task after incrementally learning up to the $t$-th task, where $j \leq t$.

The \textit{Last Accuracy} ($A_T$) is defined as the average accuracy across all tasks after completing the entire sequence of $T$ tasks:
\begin{equation}\small
A_T = \frac{1}{T} \sum_{j=1}^{T} \mathcal{A}_{T,j}.
\vspace{-3pt}
\end{equation}

The \textit{Average Accuracy} ($\overline{A}$) is defined as the average of accuracies measured immediately after learning each task:
\begin{equation}\small
\overline{A} = \frac{1}{T} \sum_{t=1}^{T} {A}_{t}.
\end{equation}

These complementary metrics provide insights into both the model's final performance stability ($A_T$) and its learning trajectory characteristics throughout the continual learning process ($\overline{A}$).
\section{Investigating the Implicit Assumptions of Continual Learning by Libcontinual}

Leveraging its unified and reproducible framework, LibContinual provides an ideal platform for the rigorous and fair evaluation of diverse continual learning algorithms. While the proliferation of CL methods has yielded impressive results on standard benchmarks, we observe that prevailing CL evaluation paradigms often rest on idealized, implicit assumptions, which can lead to an overestimation of the real-world applicability of these algorithms. In this section, we formally identify these assumptions and introduce the corresponding investigative dimensions supported by LibContinual. 
Specifically, we systematically identify and investigate the following three assumptions: (1) the availability of offline data accessibility, (2) intra-task semantic homogeneity, and (3) unregulated memory resources.

\subsection{Investigation of Assumption of Offline Data Accessibility}
The first assumption we scrutinize is the \textit{assumption of offline data accessibility}. A foundational convention in many CL evaluations is that the entire dataset $D_t$ for an incoming task is available to the learner, permitting multi-epoch training. This offline paradigm allows models to achieve thorough convergence by repeatedly optimizing over the task data before advancing to the next task.

However, this assumption contrasts with realistic scenarios, where data arrives as a single-pass stream~\cite{ocm2022, Zhuang2024}. In real-world applications, such as autonomous robots learning on the fly and edge devices processing sensor feeds, data samples are often ephemeral. In this case, they must be processed immediately and cannot be stored for repeated offline rehearsal due to strict latency constraints. This assumption is further challenged in the era of large foundation models, where growing privacy concerns and data regulations often prohibit the long-term storage of raw incoming user data, mandating a ``train-once" paradigm. The discrepancy between the offline assumption and the online reality conceals critical weaknesses in a model’s learning efficiency, specifically its ability to adapt rapidly from limited data exposure.


To address this, we introduce the Data Stream Dimension of investigation within LibContinual. Through a strict online continual learning (online CL) setting, LibContinual enables a systematic evaluation of the learning efficiency and stability across diverse methods, directly investigating the prevalent assumption of multi-epoch data access.

\subsection{Investigation of Assumption of Unregulated Memory Resources}\label{subsec:assumption_memory}

The third critical assumption we scrutinize is the \textit{Assumption of Unregulated Memory Resources}. 
In standard CL evaluations, methods are frequently compared solely on accuracy, with memory usage treated as a secondary or loosely defined constraint. This practice implicitly assumes that different forms of memory, whether raw pixels, abstract features, or model parameters, are interchangeable or negligible in cost.
However, this assumption obscures a notable flaw in comparative analysis: methods are not evaluated on an equitable basis.
First, the quantity of auxiliary memory required varies dramatically between methods~\cite{icarl2017, wa2020, l2p2022, ranpac2023}. 
More fundamentally, the \textit{qualitative nature} of the stored information is heterogeneous. 
For instance, replay methods store raw image samples~\cite{icarl2017}, while others retain abstract features~\cite{lwf2016, rapf2024}, or even task-specific parameters~\cite{api2023, dualprompt2022}. 
This qualitative diversity renders direct comparison impossible without a unified standard to account for the specific costs of these disparate storage forms.

\begin{figure}[t]
  \centering
   \includegraphics[width=\linewidth]{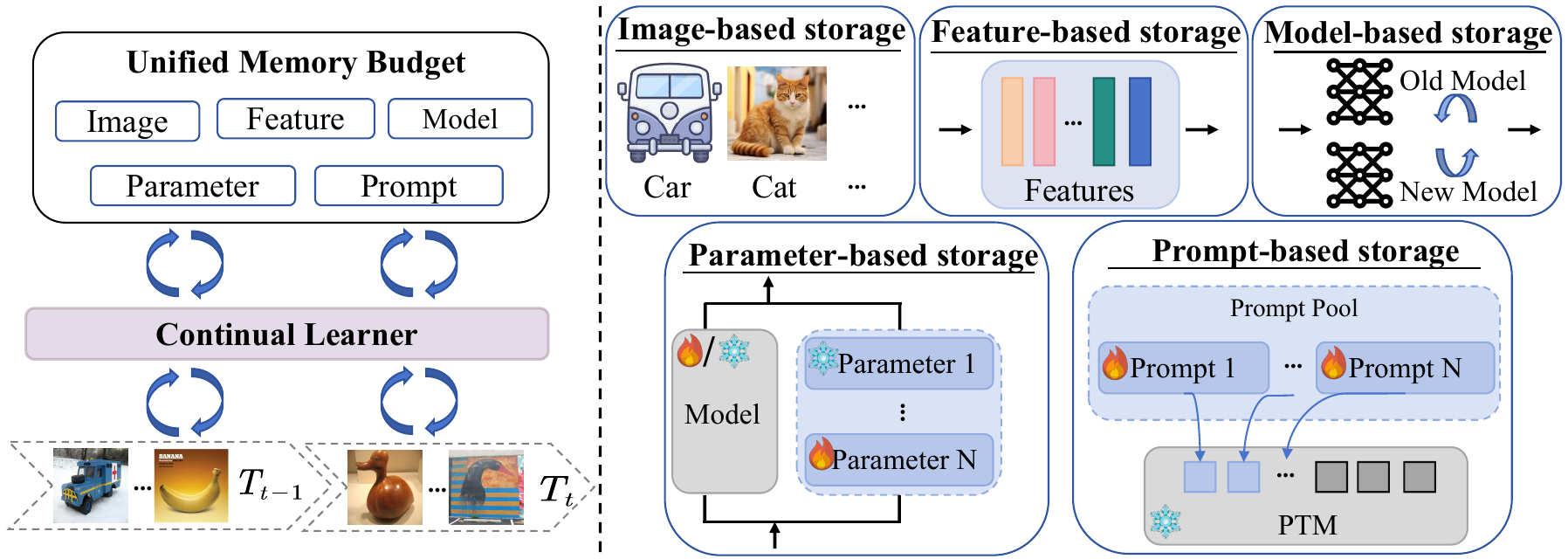}
   \caption{
   The taxonomy of continual learning methods from a storage-centric perspective. 
   The five categories, image-based, feature-based, model-based, parameter-based, and prompt-based, are illustrated with representative examples of the type of content stored in memory.
   }
   \label{fig:memory_catagory}
\vspace{-10pt}
\end{figure}

To address this challenge and enable fair comparison, we introduce the {unified memory budget} protocol within LibContinual. 
This protocol is guided by a novel storage-centric taxonomy, illustrated in Figure~\ref{fig:memory_catagory}. We strictly categorize methods based on their form of preserved knowledge into five distinct types: Image-based, Feature-based, Model-based, Parameter-based, and Prompt-based. 
Crucially, LibContinual converts these qualitatively different storage forms into a single quantitative metric: total memory usage in Megabytes (MB). 
By enforcing a strict, unified budget across all methods, we make their cost-benefit trade-offs explicit. 
This allows researchers to equitably assess whether the performance gain of a method justifies its specific memory overhead.
Below, we detail the characteristics and trade-offs of each storage category supported by our framework.

\begin{table*}[t]\small
\centering
\caption{Experimental results for the reproduction of CL methods. 
Task settings follow the format “bX-inc-task”, where “bX” denotes the total number of base classes, “inc” denotes the number of classes per incremental task, and “task” denotes the total number of tasks. “Avg.”and “Last.” refer to “Average Accuracy” and “Last Accuracy”, respectively, as defined in Section~\ref{subsection:metrics}. 
The “Reported” column shows the results from the original papers, while the “Ours” column shows the results reproduced with the LibContinual.
All methods are arranged in chronological order of their publication.}
\resizebox{\textwidth}{!}{
\setlength{\heavyrulewidth}{1.2pt}  
\setlength{\lightrulewidth}{0.2pt}  
\begin{tabular}{lcccccccc}
\toprule  
\textbf{Method} & \textbf{Backbone} & \textbf{Buffer} & \textbf{Learning Rate/Optimizer/Decay} & \textbf{Task Setting} & \textbf{TIL/CIL} & \textbf{Last./Avg. } & \textbf{Reported} & \textbf{Ours} \\
\midrule  
\multirow{2}{*}{LwF~\cite{lwf2016}} & \multirow{2}{*}{ResNet32} & \multirow{2}{*}{0} & \multirow{2}{*}{0.3/SGD/Step} & b0-10-10 & \multirow{2}{*}{CIL} & Avg. & 44.40 & 44.88 \\
 & & & & b0-20-5 & & Avg. & 54.40 & 56.38 \\
\midrule  
\multirow{2}{*}{EWC~\cite{ewc2017}} & \multirow{2}{*}{ResNet32} & \multirow{2}{*}{0} & \multirow{2}{*}{0.1/SGD/Step} & b0-10-10 & \multirow{2}{*}{CIL} & Last. & 13.10 & 10.95 \\
 & & & & b0-20-5 & & Last. & 21.90 & 20.55 \\
\midrule
\multirow{2}{*}{iCaRL~\cite{icarl2017}} & \multirow{2}{*}{ResNet32} & \multirow{2}{*}{2000} & \multirow{2}{*}{0.05/SGD/Step} & b0-10-10 & \multirow{2}{*}{CIL} & Avg. & 64.10 & 63.67 \\
 & & & & b0-20-5 & & Avg. & 67.20 & 66.83 \\
\midrule
\multirow{2}{*}{BiC~\cite{bic2019}} & \multirow{2}{*}{ResNet32} & \multirow{2}{*}{2000} & \multirow{2}{*}{0.1/SGD/Step} & b20-20-5 & \multirow{2}{*}{CIL} & Last. & 56.69 & 54.09 \\
 & & & & b50-50-2 & & Last. & 63.00 & 63.03 \\
\midrule
\multirow{2}{*}{LUCIR~\cite{lucir2019}} & \multirow{2}{*}{ResNet32} & \multirow{2}{*}{2000} & \multirow{2}{*}{0.1/SGD/Step} & b50-10-6 & \multirow{2}{*}{CIL} & Avg. & 63.42 & 62.34 \\
 & & & & b50-5-11 & & Avg. & 60.18 & 58.22 \\
\midrule
\multirow{2}{*}{WA~\cite{wa2020}} & \multirow{2}{*}{ResNet32} & \multirow{2}{*}{10000} & \multirow{2}{*}{0.1/SGD/Step} & b0-20-5 & \multirow{2}{*}{CIL} & Last. & 59.20 & 58.58 \\
 & & & & b0-10-10 & & Last. & 52.40 & 51.62 \\
\midrule
GPM~\cite{gpm2021} & AlexNet-5 & 0 & 0.01/SGD/PatienceSchedule & b10-10-10 & TIL & Last. & 72.48 & 74.43 \\
\midrule
ERAML~\cite{erace2022} & ResNet18 & 10000 & 0.1/SGD/Constant & b0-5-20 & CIL & Last. & 24.30 & 18.07 \\
\midrule
ERACE~\cite{erace2022} & ResNet18 & 10000 & 0.1/SGD/Constant & b0-5-20 & CIL & Last. & 25.80 & 26.04 \\
\midrule
TRGP~\cite{trgp2022} & AlexNet-5 & 0 & 0.01/SGD/PatienceSchedule & b0-10-10 & TIL & Last. & 74.46 & 78.22 \\
\midrule
L2P~\cite{l2p2022} & ViT-B/16 & 0 & 0.03/Adam/Constant & b0-10-10 & CIL & Last. & 83.83 & 82.85 \\
\midrule
\multirow{2}{*}{OCM~\cite{ocm2022}} & \multirow{2}{*}{ResNet18} & \multirow{2}{*}{5000} & \multirow{2}{*}{0.001/Adam/Constant} & b0-10-10 & \multirow{2}{*}{CIL} & Last. & 42.40 & 43.91 \\
 & & & & b0-2-50 & & Last. & 42.20 & 42.77 \\
\midrule
DualPrompt~\cite{dualprompt2022} & ViT-B/16 & 0 & 0.001/Adam/Cosine & b0-10-10 & CIL & Last. & 83.05 & 83.22 \\
\midrule
API~\cite{api2023} & AlexNet-5 & 0 & 0.01/SGD/PatienceSchedule & b5-5-20 & TIL & Last. & 81.40 & 80.93 \\
\midrule
CodaPrompt~\cite{codaprompt2023} & ViT-B/16 & 0 & 0.001/Adam/Step & b0-10-10 & CIL & Last. & 86.25 & 85.33 \\
\midrule
\multirow{2}{*}{RanPAC~\cite{ranpac2023}} & \multirow{2}{*}{ViT-B/16} & \multirow{2}{*}{0} & \multirow{2}{*}{0.01/SGD/Cosine} & b0-20-5 & \multirow{2}{*}{CIL} & Last. & 92.20 & 92.43 \\
 & & & & b0-10-10 & & Last. & 92.40 & 91.83 \\
\midrule
InfLoRA~\cite{inflora2024} & ViT-B/16 & 0 & 0.0005/Adam/Step & b0-10-10 & CIL & Last. & 86.51 & 86.54 \\
\midrule
MoE-Adapter4CL~\cite{moe4cl2024} & CLIP-ViT-B/16 & 0 & 0.001/AdamW/Step & b0-10-10 & CIL & Last. & 77.52 & 78.91 \\
\midrule
RAPF~\cite{rapf2024} & CLIP-ViT-B/16 & 0 & 0.001/Adam/Step & b0-10-10 & CIL & Avg. & 86.19 & 85.53 \\
\midrule
SD-LoRA~\cite{sdlora2025} & ViT-B/16 & 0 & 0.008/SGD/Constant & b0-10-10 & CIL & Avg. & 92.54 & 91.63 \\
\bottomrule  
\end{tabular}\label{tab:reproduce}
}
\vspace{-10pt}
\end{table*}

\subsubsection{Image-based storage}
Methods such as iCaRL~\cite{icarl2017}, BiC~\cite{bic2019}, LUCIR~\cite{lucir2019}, WA~\cite{wa2020}, OCM~\cite{ocm2022}, and ERAML/ERACE~\cite{erace2022} rely on storing raw input-label pairs \((x, y)\). 
Since these methods preserve the highest-fidelity representations of the original data distributions \(\mathcal{D}_t\), rehearsal is grounded in authentic samples, providing a robust defense against evaluation bias. 
However, this advantage comes with significant costs: storing raw data incurs substantial memory overhead and raises privacy concerns regarding sensitive inputs.
Furthermore, the limited buffer size creates an inherent data imbalance, often biasing models toward new classes. 
Consequently, these methods require sophisticated exemplar selection strategies, such as herding~\cite{icarl2017} or diversity maximization, to optimize the utility of the limited storage capacity.

\subsubsection{Feature-based storage}
Methods such as LwF~\cite{lwf2016}, GPM~\cite{gpm2021}, TRGP~\cite{trgp2022}, API~\cite{api2023}, InfLoRA~\cite{inflora2024}, and RAPF~\cite{rapf2024} store compressed intermediate representations or gradients to mitigate forgetting. 
Compared to raw inputs, this paradigm offers a more compact footprint and better satisfies privacy constraints by avoiding the retention of original pixels.
However, a critical limitation is the loss of representational detail.
Stored data consists of abstracted high-level semantic features. These approximations inevitably lose fine-grained information present in the source data. 
As the number of tasks increases, this information loss accumulates, potentially capping the model's final performance. 
Accordingly, recent methods focus on enhancing the fidelity of these feature representations to address this bottleneck.

\subsubsection{Model-based storage}
Methods such as EWC~\cite{ewc2017}, BIC~\cite{bic2019}, LUCIR~\cite{lucir2019}, WA~\cite{wa2020}, and CoMA~\cite{model1_2024} utilize model snapshots rather than data samples. 
This paradigm offers significant privacy advantages by strictly avoiding data retention. Furthermore, by storing complete or partial states of previous models, they provide a comprehensive reference for knowledge distillation, effectively stabilizing the learning trajectory.
However, the scalability of model storage is a major challenge. 
With the growing size of modern foundation models, storing even a single historical snapshot can be prohibitively expensive. 
Consequently, these methods face a critical trade-off between the stability provided by model snapshots and the severe constraints of storage capacity.

\subsubsection{Parameter-based storage}
Methods such as RanPAC~\cite{ranpac2023}, MoE-Adapter4CL~\cite{moe4cl2024}, and RAPF~\cite{ranpac2023} employ dynamically increasing learnable parameters. 
This involves freezing the backbone and allocating dedicated parameter blocks, such as Adapters or Experts, for new tasks.
This paradigm allows for precise control over plasticity and stability.
However, the efficiency of this expansion is difficult to manage. 
Setting appropriate parameter dimensions is non-trivial; excessively large blocks waste memory, while overly limited ones impair fitting.
Furthermore, knowledge transfer across tasks is often hindered. Since parameters are frequently isolated per task or expert, effective sharing and reuse of learned representations remain a challenge. 
This suggests that while storage-efficient, the structural utilization of these parameters in CIL and DIL settings requires further improvement.

\subsubsection{Prompt-based storage}
Methods such as L2P~\cite{l2p2022}, DualPrompt~\cite{dualprompt2022}, and CODA-Prompt~\cite{codaprompt2023} store a small number of learnable tokens (prompts) that condition a frozen PTM. 
This storage paradigm represents the extreme of efficiency. It enables rapid adaptation through lightweight updates and has shown remarkable performance in online scenarios.
However, the effectiveness is inextricably linked to the quality of the underlying PTM. 
This dependency suggests that performance gains may stem more from the frozen backbone's generality than the prompt mechanism itself. 
Additionally, finding a global optimal solution within the extremely limited parameter space of prompts remains an open optimization challenge.

\subsection{Investigation of Assumption of Intra-task Semantic Homogeneity}\label{subsec:assumption-semantic}
The second assumption is the \textit{Assumption of Intra-Task Semantic Homogeneity}. In standard benchmark constructions, whether in traditional or cross-domain settings described in Section~\ref{subsubsec:Semantic Structure}, tasks are almost invariably formed by grouping semantically related classes. For instance, a single task might consist entirely of vehicles, animals, or handwritten digits.

However, we argue that this widely accepted convention implicitly relies on a critical, unexamined assumption: Intra-Task Semantic Homogeneity.  
This design provides an implicit contextual shortcut, enabling the model to leverage task-level semantic regularities to simplify learning.
Consequently, this evaluation practice fails to distinguish if the model can learn robust, independent class representations. This convention systematically overestimates a model's true continual learning ability, as its success may hinge on exploiting these convenient but unrealistic structural regularities rather than on a genuine capacity to manage a disorganized knowledge base.

In real-world applications,  new concepts may arrive without semantic ordering. For instance, a home robot may need to learn a new plant and new shoes on the same day, driven by daily events. Similarly, a retail inventory system might process a shipment containing both new smartphones and organic snacks, grouped by logistical convenience rather than category. In these realistic scenarios, the assumption of semantic coherence breaks down.

To systematically investigate the impact of this assumption, we introduce the Semantic Structure Dimension of investigation within LibContinual. The core of this investigation is the category-randomized setting, a novel and rigorous evaluation protocol designed to strip away semantic shortcuts.

In contrast to existing setups, the category-randomized setting is defined by both low intra-task and low inter-task semantic similarity, as shown in Figure~\ref{fig:setting_datacomposition}. LibContinual implements this by first aggregating all available classes from a diverse pool of datasets (\textit{e.g.}, combining CIFAR-10, MNIST, SVHN, etc.) and then randomly shuffling them before partitioning them into tasks. This process deliberately breaks any semantic locality. As a result, a single task $\mathcal{T}_t$ becomes a semantically heterogeneous mixture, potentially containing the digit `7' alongside images of `dogs' and `airplanes'.

This challenging setup eliminates the implicit semantic context at the task level. By preventing the model from using task-level regularities as a contextual cue, the category-randomized setting forces the model to learn disparate concepts simultaneously and maintain discriminative boundaries between unrelated classes. It thereby forces the model to learn more general and robust representations for each class independently, providing a truer test of its ability to overcome catastrophic forgetting.
\section{Experimental Results by libcontinual}
\label{sec:experiment}

\subsection{Implementation Verification}

\begin{figure}[t]
  \centering
   \includegraphics[width=\linewidth]{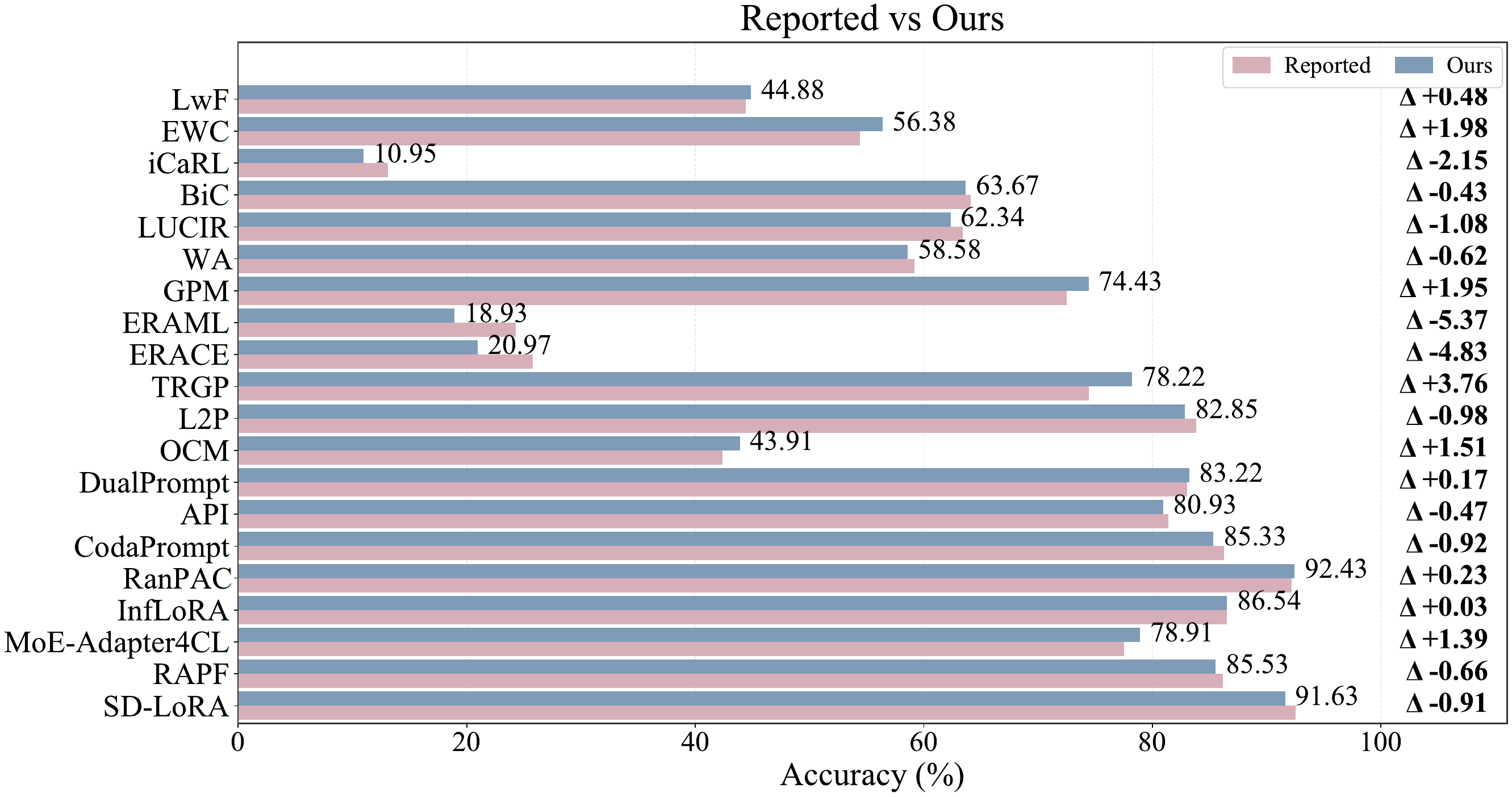}
   \caption{
   Comparison of reproduced accuracies among different continual learning methods (Reported vs Ours).
   }
   \label{fig:rep_ours}
\vspace{-15pt}
\end{figure}

To validate our re-implementation, we adopt the original settings of 19 continual learning methods (Section~\ref{subsection:implemented}) and systematically reproduce them using the unified LibContinual framework. 
Specifically, we employ the task partitioning settings as described in their respective papers, and utilize two commonly used metrics (Last accuracy and Average accuracy). 
The experiments are rigorously conducted in accordance with the backbones, buffer sizes (\textit{i.e.}, the number of stored images), learning rates, optimizers, and decay schedules specified in the original settings, and are performed under five distinct random seed configurations. As shown in Table~\ref{tab:reproduce} and Figure~\ref{fig:rep_ours}, the discrepancies between our reproduced results and the originally reported ones fall within an acceptable range. 
Specifically, for most methods, the absolute differences in their performance metrics are within ±2\%. However, some methods do not utilize multiple random seeds in their experiments, which leads to a certain degree of result variation. 
The current experimental results sufficiently confirm the accuracy of the reproduction functionality provided by LibContinual.

\begin{table*}[t]
\centering
\caption{
Comprehensive evaluation of various methods in online continual learning setting.
\textit{Base} indicates the training paradigm: \textit{PTM-based} (pre-trained model based continual learning) or \textit{Training-from-scratch} (training-from-scratch continual learning).
The best results are in \textbf{bold}, and the second-best are \underline{underlined}.
}
\label{tab:online}
\resizebox{1\textwidth}{!}{
\begin{tabular}{l|c|cc|cc|cc|cc}
\toprule
\multirow{2}{*}{Method} & \multirow{2}{*}{Base} & \multicolumn{2}{c|}{CIFAR10} & \multicolumn{2}{c|}{CIFAR100} & \multicolumn{2}{c|}{TinyImageNet} & \multicolumn{2}{c}{ImageNet-R} \\
\cmidrule{3-10}
& & Last Acc. & Avg Acc. & Last Acc. & Avg Acc. & Last Acc. & Avg Acc. & Last Acc. & Avg Acc. \\
\midrule
LwF~\cite{lwf2016} & Training-from-scratch & 
26.60 \textcolor{gray}{\scriptsize$\pm$ 1.25} & 
46.54 \textcolor{gray}{\scriptsize$\pm$ 4.52} & 
9.22 \textcolor{gray}{\scriptsize$\pm$ 0.73} & 
13.53 \textcolor{gray}{\scriptsize$\pm$ 0.94} & 
7.49 \textcolor{gray}{\scriptsize$\pm$ 0.48} & 
11.05 \textcolor{gray}{\scriptsize$\pm$ 0.62} & 
2.38 \textcolor{gray}{\scriptsize$\pm$ 0.38} & 
4.06 \textcolor{gray}{\scriptsize$\pm$ 0.64} \\
EWC~\cite{ewc2017} & Training-from-scratch & 
10.00 \textcolor{gray}{\scriptsize$\pm$ 0.00} & 
35.91 \textcolor{gray}{\scriptsize$\pm$ 3.69} & 
5.27 \textcolor{gray}{\scriptsize$\pm$ 0.45} & 
11.51 \textcolor{gray}{\scriptsize$\pm$ 1.20} & 
5.00 \textcolor{gray}{\scriptsize$\pm$ 0.77} & 
9.37 \textcolor{gray}{\scriptsize$\pm$ 0.90} & 
1.52 \textcolor{gray}{\scriptsize$\pm$ 0.14} & 
2.81 \textcolor{gray}{\scriptsize$\pm$ 0.52} \\
iCaRL~\cite{icarl2017} & Training-from-scratch & 
42.78 \textcolor{gray}{\scriptsize$\pm$ 2.43} & 
55.71 \textcolor{gray}{\scriptsize$\pm$ 6.45} & 
16.16 \textcolor{gray}{\scriptsize$\pm$ 1.40} & 
19.74 \textcolor{gray}{\scriptsize$\pm$ 1.64} & 
10.54 \textcolor{gray}{\scriptsize$\pm$ 0.88} & 
12.87 \textcolor{gray}{\scriptsize$\pm$ 1.78} & 
2.98 \textcolor{gray}{\scriptsize$\pm$ 0.98} & 
4.83 \textcolor{gray}{\scriptsize$\pm$ 1.12} \\
BiC~\cite{bic2019} & Training-from-scratch & 
27.62 \textcolor{gray}{\scriptsize$\pm$ 3.44} & 
45.79 \textcolor{gray}{\scriptsize$\pm$ 2.84} & 
5.97 \textcolor{gray}{\scriptsize$\pm$ 0.66} & 
13.00 \textcolor{gray}{\scriptsize$\pm$ 0.86} & 
2.23 \textcolor{gray}{\scriptsize$\pm$ 0.26} & 
4.85 \textcolor{gray}{\scriptsize$\pm$ 0.72} & 
1.07 \textcolor{gray}{\scriptsize$\pm$ 0.47} & 
3.48 \textcolor{gray}{\scriptsize$\pm$ 0.31} \\
LUCIR~\cite{lucir2019} & Training-from-scratch & 
23.63 \textcolor{gray}{\scriptsize$\pm$ 2.59} & 
45.02 \textcolor{gray}{\scriptsize$\pm$ 2.71} & 
6.51 \textcolor{gray}{\scriptsize$\pm$ 0.64} & 
12.81 \textcolor{gray}{\scriptsize$\pm$ 0.88} & 
3.10 \textcolor{gray}{\scriptsize$\pm$ 0.44} & 
6.94 \textcolor{gray}{\scriptsize$\pm$ 1.00} & 
1.48 \textcolor{gray}{\scriptsize$\pm$ 0.35} & 
3.92 \textcolor{gray}{\scriptsize$\pm$ 0.50} \\
WA~\cite{wa2020} & Training-from-scratch & 
37.18 \textcolor{gray}{\scriptsize$\pm$ 5.67} & 
52.94 \textcolor{gray}{\scriptsize$\pm$ 2.42} & 
9.52 \textcolor{gray}{\scriptsize$\pm$ 1.04} & 
18.61 \textcolor{gray}{\scriptsize$\pm$ 1.03} & 
3.99 \textcolor{gray}{\scriptsize$\pm$ 0.58} & 
9.17 \textcolor{gray}{\scriptsize$\pm$ 0.35} & 
2.62 \textcolor{gray}{\scriptsize$\pm$ 0.26} & 
5.98 \textcolor{gray}{\scriptsize$\pm$ 0.60} \\
GPM~\cite{gpm2021} & Training-from-scratch & 
29.24 \textcolor{gray}{\scriptsize$\pm$ 0.78} & 
47.00 \textcolor{gray}{\scriptsize$\pm$ 2.57} & 
13.40 \textcolor{gray}{\scriptsize$\pm$ 0.23} & 
22.49 \textcolor{gray}{\scriptsize$\pm$ 1.14} & 
2.41 \textcolor{gray}{\scriptsize$\pm$ 0.12} & 
5.13 \textcolor{gray}{\scriptsize$\pm$ 0.33} & 
2.77 \textcolor{gray}{\scriptsize$\pm$ 0.42} & 
7.36 \textcolor{gray}{\scriptsize$\pm$ 0.84} \\
ERAML~\cite{erace2022} & Training-from-scratch & 
50.82 \textcolor{gray}{\scriptsize$\pm$ 1.99} & 
65.64 \textcolor{gray}{\scriptsize$\pm$ 1.81} & 
21.67 \textcolor{gray}{\scriptsize$\pm$ 0.56} & 
28.07 \textcolor{gray}{\scriptsize$\pm$ 0.77} & 
15.29 \textcolor{gray}{\scriptsize$\pm$ 0.92} & 
20.32 \textcolor{gray}{\scriptsize$\pm$ 0.54} & 
4.75 \textcolor{gray}{\scriptsize$\pm$ 0.86} & 
7.14 \textcolor{gray}{\scriptsize$\pm$ 0.80} \\
ERACE~\cite{erace2022} & Training-from-scratch & 
48.35 \textcolor{gray}{\scriptsize$\pm$ 5.70} & 
65.57 \textcolor{gray}{\scriptsize$\pm$ 1.92} & 
24.71 \textcolor{gray}{\scriptsize$\pm$ 0.21} & 
32.42 \textcolor{gray}{\scriptsize$\pm$ 0.57} & 
20.62 \textcolor{gray}{\scriptsize$\pm$ 1.09} & 
26.53 \textcolor{gray}{\scriptsize$\pm$ 1.17} & 
7.39 \textcolor{gray}{\scriptsize$\pm$ 0.79} & 
11.03 \textcolor{gray}{\scriptsize$\pm$ 0.85} \\
TRGP~\cite{trgp2022} & Training-from-scratch & 
27.06 \textcolor{gray}{\scriptsize$\pm$ 0.34} & 
46.73 \textcolor{gray}{\scriptsize$\pm$ 2.56} & 
13.36 \textcolor{gray}{\scriptsize$\pm$ 0.40} & 
23.01 \textcolor{gray}{\scriptsize$\pm$ 1.29} & 
2.09 \textcolor{gray}{\scriptsize$\pm$ 0.13} & 
5.00 \textcolor{gray}{\scriptsize$\pm$ 0.22} & 
3.08 \textcolor{gray}{\scriptsize$\pm$ 0.48} & 
7.45 \textcolor{gray}{\scriptsize$\pm$ 1.00} \\
API~\cite{api2023} & Training-from-scratch & 
26.66 \textcolor{gray}{\scriptsize$\pm$ 1.57} & 
44.44 \textcolor{gray}{\scriptsize$\pm$ 2.89} & 
12.85 \textcolor{gray}{\scriptsize$\pm$ 0.35} & 
26.27 \textcolor{gray}{\scriptsize$\pm$ 0.54} & 
1.67 \textcolor{gray}{\scriptsize$\pm$ 0.05} & 
5.11 \textcolor{gray}{\scriptsize$\pm$ 0.11} & 
1.84 \textcolor{gray}{\scriptsize$\pm$ 0.33} & 
4.56 \textcolor{gray}{\scriptsize$\pm$ 0.95} \\
L2P~\cite{l2p2022} & PTM-based & 
87.61 \textcolor{gray}{\scriptsize$\pm$ 3.51} & 
93.70 \textcolor{gray}{\scriptsize$\pm$ 1.05} & 
79.07 \textcolor{gray}{\scriptsize$\pm$ 1.22} & 
84.68 \textcolor{gray}{\scriptsize$\pm$ 0.99} & 
82.47 \textcolor{gray}{\scriptsize$\pm$ 0.43} & 
87.09 \textcolor{gray}{\scriptsize$\pm$ 0.28} & 
62.78 \textcolor{gray}{\scriptsize$\pm$ 0.93} & 
67.42 \textcolor{gray}{\scriptsize$\pm$ 1.59} \\
OCM~\cite{ocm2022} & PTM-based & 
77.91 \textcolor{gray}{\scriptsize$\pm$ 2.15} & 
82.36 \textcolor{gray}{\scriptsize$\pm$ 1.59} & 
41.20 \textcolor{gray}{\scriptsize$\pm$ 0.81} & 
44.00 \textcolor{gray}{\scriptsize$\pm$ 1.69} & 
20.04 \textcolor{gray}{\scriptsize$\pm$ 0.32} & 
24.80 \textcolor{gray}{\scriptsize$\pm$ 0.84} & 
2.12 \textcolor{gray}{\scriptsize$\pm$ 0.41} & 
4.27 \textcolor{gray}{\scriptsize$\pm$ 0.50} \\
DualPrompt~\cite{dualprompt2022} & PTM-based & 
82.34 \textcolor{gray}{\scriptsize$\pm$ 1.85} & 
91.00 \textcolor{gray}{\scriptsize$\pm$ 1.15} & 
76.22 \textcolor{gray}{\scriptsize$\pm$ 0.41} & 
83.28 \textcolor{gray}{\scriptsize$\pm$ 0.72} & 
81.21 \textcolor{gray}{\scriptsize$\pm$ 0.29} & 
86.06 \textcolor{gray}{\scriptsize$\pm$ 0.76} & 
62.76 \textcolor{gray}{\scriptsize$\pm$ 0.59} & 
69.45 \textcolor{gray}{\scriptsize$\pm$ 1.54} \\
CodaPrompt~\cite{codaprompt2023} & PTM-based & 
84.48 \textcolor{gray}{\scriptsize$\pm$ 3.30} & 
91.23 \textcolor{gray}{\scriptsize$\pm$ 2.62} & 
81.73 \textcolor{gray}{\scriptsize$\pm$ 0.14} & 
87.17 \textcolor{gray}{\scriptsize$\pm$ 0.70} & 
\underline{84.42} \textcolor{gray}{\scriptsize$\pm$ 0.40} & 
{88.89} \textcolor{gray}{\scriptsize$\pm$ 0.41} & 
68.87 \textcolor{gray}{\scriptsize$\pm$ 0.83} & 
75.14 \textcolor{gray}{\scriptsize$\pm$ 0.49} \\
RanPAC~\cite{ranpac2023} & PTM-based & 
\textbf{94.42} \textcolor{gray}{\scriptsize$\pm$ 2.61} & 
\textbf{98.10} \textcolor{gray}{\scriptsize$\pm$ 0.71} & 
\textbf{88.14} \textcolor{gray}{\scriptsize$\pm$ 0.84} & 
\textbf{93.27} \textcolor{gray}{\scriptsize$\pm$ 0.57} & 
{84.09} \textcolor{gray}{\scriptsize$\pm$ 0.43} & 
\underline{90.33} \textcolor{gray}{\scriptsize$\pm$ 0.23} & 
69.70 \textcolor{gray}{\scriptsize$\pm$ 0.48} & 
76.10 \textcolor{gray}{\scriptsize$\pm$ 1.16} \\
InfLoRA~\cite{inflora2024} & PTM-based & 
86.72 \textcolor{gray}{\scriptsize$\pm$ 5.73} & 
92.32 \textcolor{gray}{\scriptsize$\pm$ 3.05} & 
\underline{82.96} \textcolor{gray}{\scriptsize$\pm$ 0.58} & 
\underline{88.99} \textcolor{gray}{\scriptsize$\pm$ 0.31} & 
80.33 \textcolor{gray}{\scriptsize$\pm$ 0.49} & 
86.96 \textcolor{gray}{\scriptsize$\pm$ 0.42} & 
67.36 \textcolor{gray}{\scriptsize$\pm$ 1.34} & 
73.63 \textcolor{gray}{\scriptsize$\pm$ 2.32} \\
MoE-Adapter4CL~\cite{moe4cl2024} & PTM-based & 
89.21 \textcolor{gray}{\scriptsize$\pm$ 5.35} & 
94.34 \textcolor{gray}{\scriptsize$\pm$ 1.49} & 
79.42 \textcolor{gray}{\scriptsize$\pm$ 0.21} & 
86.03 \textcolor{gray}{\scriptsize$\pm$ 0.43} & 
76.68 \textcolor{gray}{\scriptsize$\pm$ 0.08} & 
83.71 \textcolor{gray}{\scriptsize$\pm$ 0.43} & 
\textbf{86.59} \textcolor{gray}{\scriptsize$\pm$ 1.11} & 
\textbf{90.55} \textcolor{gray}{\scriptsize$\pm$ 1.07} \\
RAPF~\cite{rapf2024} & PTM-based & 
\underline{94.39} \textcolor{gray}{\scriptsize$\pm$ 0.36} & 
\underline{96.62} \textcolor{gray}{\scriptsize$\pm$ 0.61} & 
71.54 \textcolor{gray}{\scriptsize$\pm$ 1.55} & 
80.74 \textcolor{gray}{\scriptsize$\pm$ 0.97} & 
{81.65} \textcolor{gray}{\scriptsize$\pm$ 0.92} & 
{85.92} \textcolor{gray}{\scriptsize$\pm$ 0.59} & 
\underline{79.52} \textcolor{gray}{\scriptsize$\pm$ 0.29} & 
\underline{84.08} \textcolor{gray}{\scriptsize$\pm$ 0.43} \\
SD-LoRA~\cite{sdlora2025} & PTM-based & 
{89.16} \textcolor{gray}{\scriptsize$\pm$ 1.15} & 
{94.37} \textcolor{gray}{\scriptsize$\pm$ 0.48} & 
82.78 \textcolor{gray}{\scriptsize$\pm$ 0.55} & 
88.73 \textcolor{gray}{\scriptsize$\pm$ 0.31} & 
\textbf{88.72} \textcolor{gray}{\scriptsize$\pm$ 0.39} & 
\textbf{92.67} \textcolor{gray}{\scriptsize$\pm$ 0.62} & 
{75.34} \textcolor{gray}{\scriptsize$\pm$ 0.57} & 
{78.32} \textcolor{gray}{\scriptsize$\pm$ 2.81} \\
\bottomrule
\end{tabular}
}
\vspace{-10pt}
\end{table*}

\subsection{Investigation 1: The Impact of Offline Data Accessibility in Online CL}\label{subsec:online experiment}

In applications requiring rapid adaptation to streaming data (\textit{e.g.}, robot perception, real-time recommendation systems), models must continuously learn under stringent constraints. 
Specifically, prevailing online CL methods~\cite{ocm2022}~\cite{erace2022} simulate real-world environments using settings with epoch=1 and batchsize=10, and LibContinual also adopts such settings. 
Experiments (Table \ref{tab:online}) reveal a significant performance divergence between training-from-scratch methods and PTM-based approaches, motivating the following analysis.

\subsubsection{Training-from-scratch methods struggle in online CL}
Traditional continual learning methods, which are typically designed under the assumption of multi-epoch training for stable convergence, exhibit catastrophic performance degradation in online settings where data is presented in a single pass. As evidenced in Table \ref{tab:online}, these methods suffer from severe accuracy collapse, often approaching near-random performance levels. For instance, on CIFAR-10, EWC achieves only 10\% accuracy, while BiC drops to a mere 2.23\% on TinyImageNet. 
This failure is primarily attributed to inadequate model fitting, as these methods lack the opportunity for repeated data exposure. 
Consequently, they are unable to sufficiently optimize their complex parameter sets, leading to rapid forgetting and an inability to assimilate new knowledge effectively.
In contrast, methods with a frozen pretrained backbone like L2P perform markedly better. 
They operate within compact and efficient parameter subspaces enabled by prompt tuning or random projection layers. 
This design facilitates rapid convergence and strong results even under the strict single-epoch training constraint.
This significant performance gap underscores the dominant role of pre-trained representations in ensuring online learning efficiency, effectively overshadowing algorithmic optimizations tailored for training-from-scratch methods.

\subsubsection{Need for more challenging benchmarks}
While PTM-based methods consistently achieve high performance in online continual learning, their results on conventional benchmarks such as CIFAR-100 and TinyImageNet tend to be highly homogeneous, thereby limiting the ability to discern nuanced differences in adaptation efficiency. As illustrated in Table \ref{tab:online}, top-performing methods often cluster within a narrow performance band, for example, DualPrompt (76.22\%) and MoE-Adapter4CL (79.42\%) on CIFAR-100, making it difficult to evaluate relative strengths in model plasticity or stability. 
In contrast, more complex and semantically diverse datasets like ImageNet-R reveal significant variations in capability, particularly among ViT-based models. For instance, while MoE-Adapter4CL attains 86.59\% on ImageNet-R, DualPrompt achieves only 62.76\%, highlighting critical differences in robustness and representational flexibility. Meanwhile, training-from-scratch methods universally collapse to accuracies below 8\% on this challenging benchmark, further exposing their limitations. These observations suggest that the current community reliance on relatively simple datasets such as CIFAR-100 or TinyImageNet is insufficient for driving meaningful progress in online continual learning. We therefore advocate for the adoption of richer, more complex benchmarks like ImageNet-R to better expose methodological distinctions and guide the development of more adaptive and scalable ViT-focused online CL algorithms.


\subsection{Investigation 2: Algorithmic Efficiency under a Unified Memory Budget}~\label{subsec:memory_experiment}
A critical limitation in current continual learning research is the lack of a standardized framework for evaluating memory overhead, which often makes direct comparisons of reported accuracies misleading. 
To address this, we conduct a rigorous, storage-centric analysis through the proposed LibContinual toolbox. We recognize that the total memory footprint of a continual learning system is composed of two main parts: the static memory for the network backbone and the dynamic, additional storage required by the specific CL strategy. 
Since different approaches may employ backbones of varying sizes (\textit{e.g.}, ResNet~\cite{resnet2016} vs. ViT~\cite{vit2021}), their static memory costs differ, making total footprint comparisons inequitable.

For a fair and direct comparison of algorithmic efficiency, we introduce the {unified memory budget}. 
This budget is defined as the total additional memory, measured in megabytes (MB), that an algorithm requires {beyond} the backbone's own parameters. 
Detailed methodology regarding memory calculation can be found in the Appendix C.
This meticulous accounting ensures a fair comparison by encompassing all sources of extra memory cost, including: 1) \textit{Image-based} storage for raw data exemplars; 2) \textit{Feature-based} storage for intermediate representations or gradients; 3) \textit{Model-based} storage for snapshots of past models needed for regularization or distillation; 4) \textit{Parameter-based} storage for dynamically expanded network modules such as adapters or new projection layers; and 5) \textit{Prompt-based} storage for learnable prompt tokens. 
By unifying these qualitatively distinct forms of memory under a single quantitative metric, we can set uniform memory budgets to observe how the performance of different methods changes as storage increases.

\begin{figure*}
    \centering
    \begin{subfigure}[b]{0.95\linewidth}\label{fig:memory_cifar10}
        \centering
        \includegraphics[width=1\linewidth]{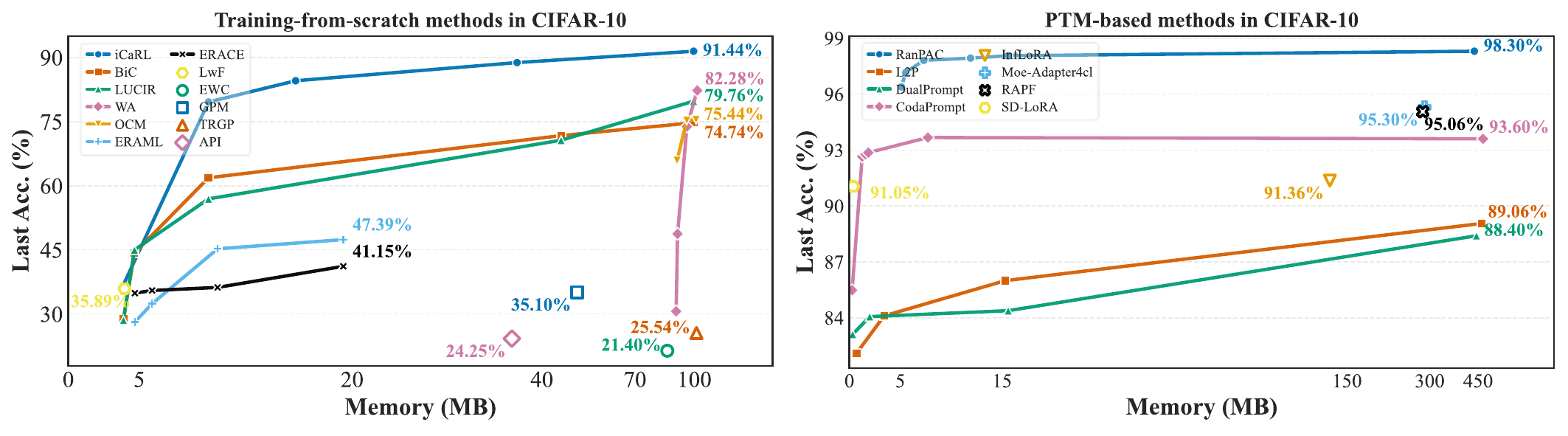}
        \caption{Comparison of last accuracy for training-from-scratch methods and PTM-based methods on different memory configurations on CIFAR-10.}
    \end{subfigure}
    
    
    \begin{subfigure}[b]{0.95\linewidth}
        \centering
        \includegraphics[width=1\linewidth]{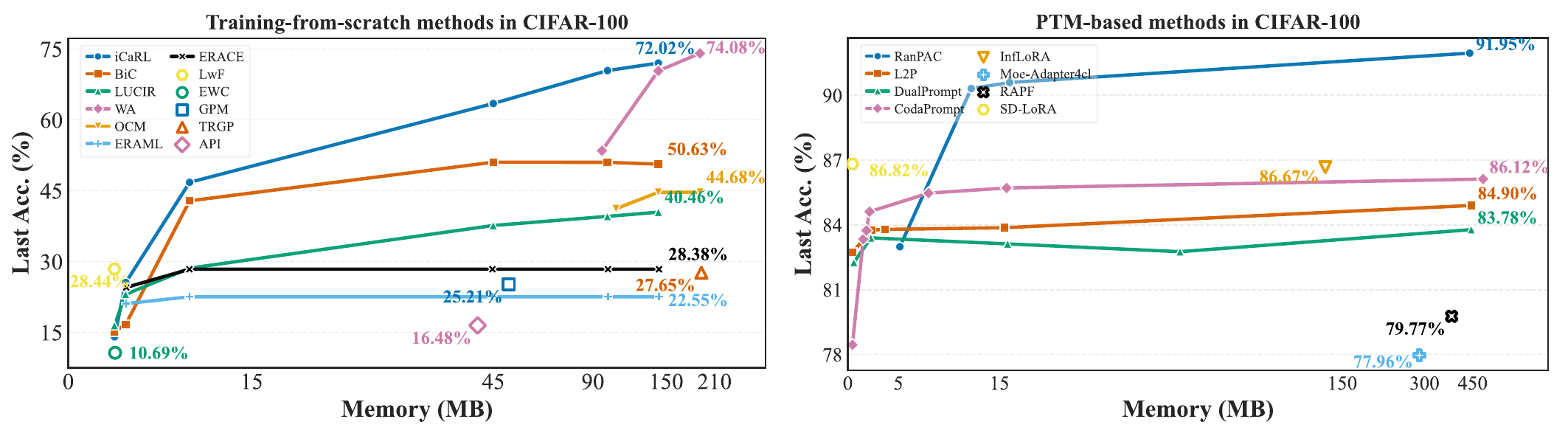}
        \caption{Comparison of last accuracy for training-from-scratch methods and PTM-based methods on different memory configurations on CIFAR-100.}
    \end{subfigure}
    
    
    \begin{subfigure}[b]{0.95\linewidth}
        \centering
        \includegraphics[width=1\linewidth]{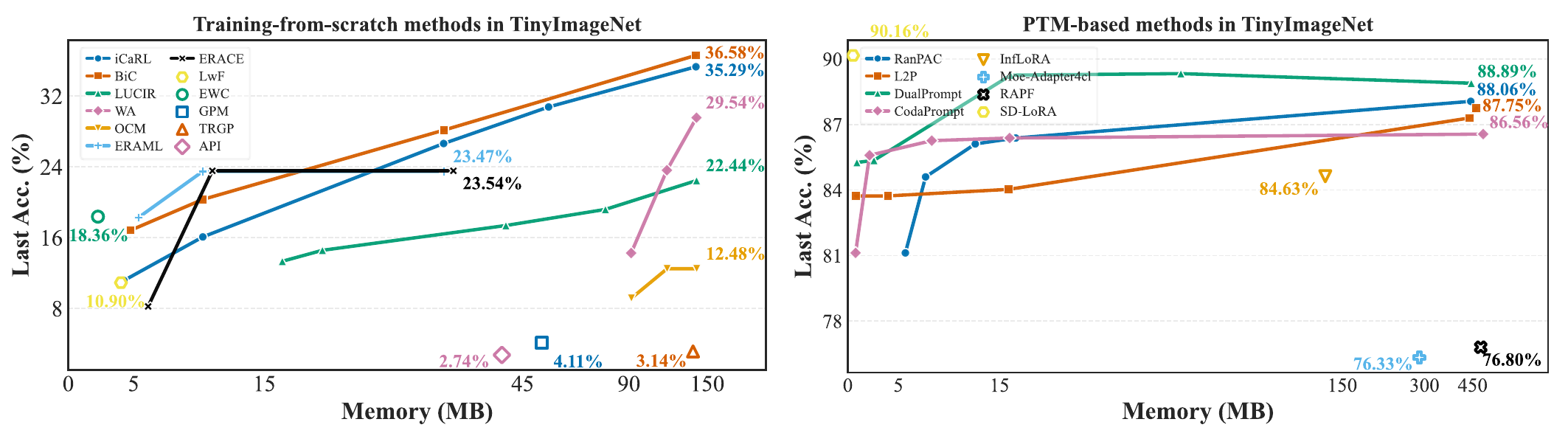}
        \caption{Comparison of last accuracy for training-from-scratch methods and PTM-based methods on different memory configurations on TinyImageNet.}
    \end{subfigure}
    
    
    \begin{subfigure}[b]{0.95\linewidth}
        \centering
        \includegraphics[width=1\linewidth]{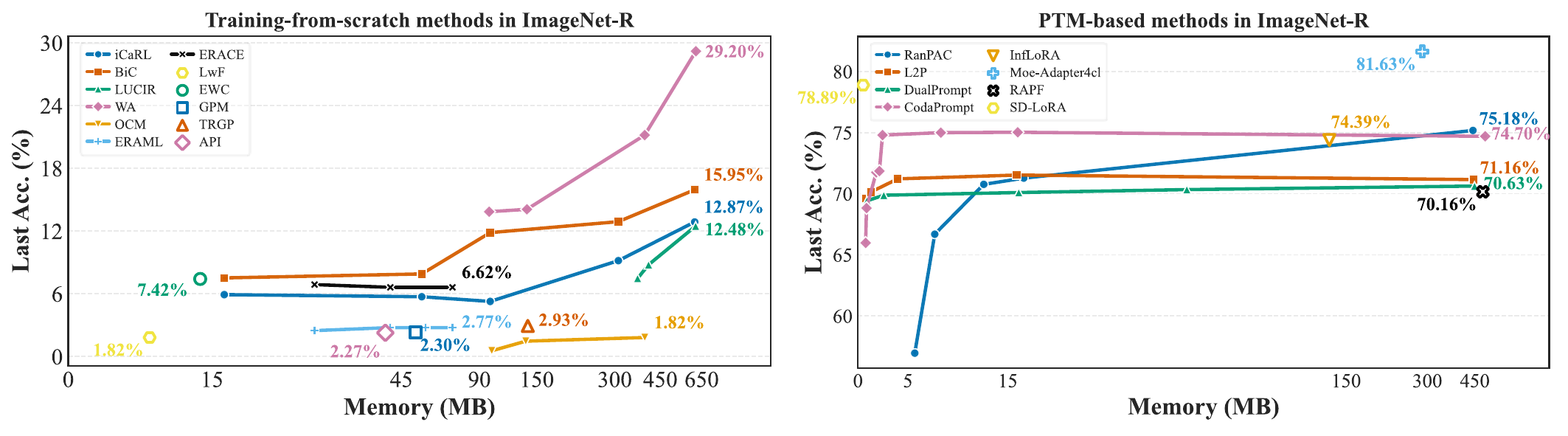}
        \caption{Comparison of last accuracy for training-from-scratch methods and PTM-based methods on different memory configurations on ImageNet-R.}
    \end{subfigure}
    
    \caption{Comparison of last accuracy achieved by the training-from-scratch methods and PTM-based methods across different memory configurations on various datasets.}
    \label{fig:scratch_memory_accuracy}
\end{figure*}

As illustrated in Figure~\ref{fig:scratch_memory_accuracy}, our experiments are conducted on four benchmark datasets: CIFAR-10, CIFAR-100, TinyImageNet, and ImageNet-R. Methods are categorized into two groups based on their training paradigm: those trained from scratch and those leveraging pre-trained models.
Notably, certain methods incapable of scaling their memory usage are represented as fixed points in the plots, reflecting their static memory allocation.

\subsubsection{Analysis of training-from-scratch methods}
For methods trained from scratch, the experimental plots reveal a clear, though often inefficient, positive correlation between memory consumption and last accuracy. 
Replay-based methods, such as iCaRL, LUCIR, and ERACE, consistently demonstrate that increasing the memory allocated for an image buffer leads to significant performance gains. 
On CIFAR-10, for instance, the accuracy of iCaRL improves dramatically from approximately 35\% to over 90\% as the memory budget expands from 4 MB to 100 MB, highlighting the substantial benefit of rehearsing past data. 
However, this strategy exhibits inefficiency and diminishing returns. 
A particularly illustrative case is WA, which consumes nearly 100 MB of memory but achieves substantially lower final accuracy (82.28\%) compared to iCaRL (91.44\%) at a comparable 100 MB budget, demonstrating iCaRL's superior memory efficiency.

In contrast, low-memory strategies that rely on model regularization, such as EWC and LwF, consistently occupy the low-performance corner of the plots. Their inability to effectively combat catastrophic forgetting on more challenging benchmarks like TinyImageNet and ImageNet-R underscores their limitations when past data is not accessible. In essence, for scratch-based methods, performance is predominantly dictated by the size of the replay buffer. This “brute-force” approach, while intuitive, is not a scalable or memory-efficient solution for realistic, resource-constrained lifelong learning scenarios.

\begin{table*}[t]\small
\centering
\caption{Cross-domain and category-randomized continual learning results on 5-datasets benchmark. The `Type' column indicates the core algorithmic strategy. In the `Difference' columns, significant gains ($\ge$8) are in \textcolor{blue}{blue}, significant losses ($\ge$8) are in \textcolor{red}{red}, and minor changes ($<$8) are in \textcolor{darkgreen}{green}.}
\label{tab:cross_domain_results}
\resizebox{\textwidth}{!}{%
\begin{tabular}{l|c|cc|cc|cc}
\toprule
\multirow{2}{*}{Method} & \multirow{2}{*}{Type} & \multicolumn{2}{c|}{Cross-Domain Setting} & \multicolumn{2}{c|}{Category-Randomized Setting} & \multicolumn{2}{c}{Difference (\textit{Cat-R} - \textit{Cross-D})} \\
\cmidrule{3-8}
& & Last Acc. & Avg Acc. & Last Acc. & Avg Acc. & $\Delta$ Last Acc. & $\Delta$ Avg Acc. \\
\midrule
LwF~\cite{lwf2016} 
& Regularization
& 39.79 \textcolor{gray}{\scriptsize$\pm$ 4.55}
& 64.23 \textcolor{gray}{\scriptsize$\pm$ 1.33}
& 54.86 \textcolor{gray}{\scriptsize$\pm$ 5.45}
& 74.12 \textcolor{gray}{\scriptsize$\pm$ 1.27}
& \textcolor{blue}{+15.07} & \textcolor{blue}{+9.89} \\
EWC~\cite{ewc2017}
& Regularization
& 26.24 \textcolor{gray}{\scriptsize$\pm$ 4.93}
& 53.24 \textcolor{gray}{\scriptsize$\pm$ 1.61} 
& 28.70 \textcolor{gray}{\scriptsize$\pm$ 5.88}
& 49.46 \textcolor{gray}{\scriptsize$\pm$ 2.42}
& \textcolor{darkgreen}{+2.46} & \textcolor{darkgreen}{-3.78} \\
iCaRL~\cite{icarl2017}
& Replay
& 81.18 \textcolor{gray}{\scriptsize$\pm$ 0.88}
& 85.70 \textcolor{gray}{\scriptsize$\pm$ 0.29}
& 81.79 \textcolor{gray}{\scriptsize$\pm$ 1.66}
& 90.32 \textcolor{gray}{\scriptsize$\pm$ 0.60}
& \textcolor{darkgreen}{+0.61} & \textcolor{darkgreen}{+4.62} \\
BiC~\cite{bic2019}
& Replay
& 51.64 \textcolor{gray}{\scriptsize$\pm$ 0.75}
& 66.98 \textcolor{gray}{\scriptsize$\pm$ 0.98}
& 47.54 \textcolor{gray}{\scriptsize$\pm$ 10.34}
& 72.19 \textcolor{gray}{\scriptsize$\pm$ 7.97}
& \textcolor{darkgreen}{-4.10} & \textcolor{darkgreen}{+5.21} \\
LUCIR~\cite{lucir2019} 
& Replay
& 66.69 \textcolor{gray}{\scriptsize$\pm$ 0.67} 
& 76.94 \textcolor{gray}{\scriptsize$\pm$ 0.92}
& 49.51 \textcolor{gray}{\scriptsize$\pm$ 6.03}
& 74.75 \textcolor{gray}{\scriptsize$\pm$ 2.90}
& \textcolor{red}{-17.18} & \textcolor{darkgreen}{-2.19} \\
WA~\cite{wa2020} 
& Replay
& 83.56 \textcolor{gray}{\scriptsize$\pm$ 0.40} 
& 87.64 \textcolor{gray}{\scriptsize$\pm$ 0.22} 
& 71.51 \textcolor{gray}{\scriptsize$\pm$ 4.22} 
& 85.33 \textcolor{gray}{\scriptsize$\pm$ 1.67} 
& \textcolor{red}{-12.05} & \textcolor{darkgreen}{-2.31} \\
GPM~\cite{gpm2021} 
& Optimization
& 69.87 \textcolor{gray}{\scriptsize$\pm$ 1.09}
& 81.08 \textcolor{gray}{\scriptsize$\pm$ 0.37}
& 62.51 \textcolor{gray}{\scriptsize$\pm$ 4.59}
& 76.43 \textcolor{gray}{\scriptsize$\pm$ 4.52}
& \textcolor{darkgreen}{-7.36} & \textcolor{darkgreen}{-4.65} \\
ERAML~\cite{erace2022} 
& Replay
& 84.11 \textcolor{gray}{\scriptsize$\pm$ 0.38}
& 80.22 \textcolor{gray}{\scriptsize$\pm$ 0.75}
& 80.12 \textcolor{gray}{\scriptsize$\pm$ 0.96}
& 86.94 \textcolor{gray}{\scriptsize$\pm$ 2.06}
& \textcolor{darkgreen}{-3.99} & \textcolor{darkgreen}{+6.72} \\
ERACE~\cite{erace2022} 
& Replay
& 85.85 \textcolor{gray}{\scriptsize$\pm$ 0.45}
& 81.76 \textcolor{gray}{\scriptsize$\pm$ 0.64}
& 82.69 \textcolor{gray}{\scriptsize$\pm$ 0.53} 
& 88.47 \textcolor{gray}{\scriptsize$\pm$ 1.32} 
& \textcolor{darkgreen}{-3.16} & \textcolor{darkgreen}{+6.71} \\
TRGP~\cite{trgp2022} 
& Optimization
& 60.43 \textcolor{gray}{\scriptsize$\pm$ 0.78}
& 72.36 \textcolor{gray}{\scriptsize$\pm$ 0.37}
& 54.20 \textcolor{gray}{\scriptsize$\pm$ 4.44}
& 72.51 \textcolor{gray}{\scriptsize$\pm$ 3.49}
& \textcolor{darkgreen}{-6.23} & \textcolor{darkgreen}{+0.15} \\
L2P~\cite{l2p2022} 
& Representation
& 64.69 \textcolor{gray}{\scriptsize$\pm$ 0.68}
& 82.69 \textcolor{gray}{\scriptsize$\pm$ 0.68}
& 47.59 \textcolor{gray}{\scriptsize$\pm$ 1.86}
& 67.39 \textcolor{gray}{\scriptsize$\pm$ 5.72}
& \textcolor{red}{-17.10} & \textcolor{red}{-15.30} \\
OCM~\cite{ocm2022}
& Replay
& 83.72 \textcolor{gray}{\scriptsize$\pm$ 0.64} 
& 78.49 \textcolor{gray}{\scriptsize$\pm$ 0.46}  
& 81.40 \textcolor{gray}{\scriptsize$\pm$ 3.44} 
& 87.77 \textcolor{gray}{\scriptsize$\pm$ 1.11}
& \textcolor{darkgreen}{-2.32} & \textcolor{blue}{+9.28} \\
DualPrompt~\cite{dualprompt2022} 
& Representation
& 73.53 \textcolor{gray}{\scriptsize$\pm$ 1.52}
& 87.05 \textcolor{gray}{\scriptsize$\pm$ 1.01}
& 52.81 \textcolor{gray}{\scriptsize$\pm$ 2.03}
& 71.21 \textcolor{gray}{\scriptsize$\pm$ 5.12}
& \textcolor{red}{-20.72} & \textcolor{red}{-15.84} \\
API~\cite{api2023} 
& Architecture
& 62.85 \textcolor{gray}{\scriptsize$\pm$ 3.55}
& 77.89 \textcolor{gray}{\scriptsize$\pm$ 2.63}
& 62.48 \textcolor{gray}{\scriptsize$\pm$ 1.88}
& 77.02 \textcolor{gray}{\scriptsize$\pm$ 2.29}
& \textcolor{darkgreen}{-0.37} & \textcolor{darkgreen}{-0.87}\\
CodaPrompt~\cite{codaprompt2023} 
& Representation
& 71.54 \textcolor{gray}{\scriptsize$\pm$ 1.87}
& 86.99 \textcolor{gray}{\scriptsize$\pm$ 0.99}
& 55.35 \textcolor{gray}{\scriptsize$\pm$ 5.59}
& 75.94 \textcolor{gray}{\scriptsize$\pm$ 3.57} 
& \textcolor{red}{-16.19} & \textcolor{red}{-11.05}\\
RanPAC~\cite{ranpac2023} 
& Representation
& 87.10 \textcolor{gray}{\scriptsize$\pm$ 0.07} 
& 94.33 \textcolor{gray}{\scriptsize$\pm$ 0.02}
& 58.02 \textcolor{gray}{\scriptsize$\pm$ 4.16}
& 77.05 \textcolor{gray}{\scriptsize$\pm$ 2.71}
& \textcolor{red}{-29.08} & \textcolor{red}{-17.28} \\
InfLoRA~\cite{inflora2024} 
& Architecture
& 83.67 \textcolor{gray}{\scriptsize$\pm$ 0.93}
& 92.78 \textcolor{gray}{\scriptsize$\pm$ 0.28}
& 57.13 \textcolor{gray}{\scriptsize$\pm$ 9.25}
& 77.41 \textcolor{gray}{\scriptsize$\pm$ 4.29}
& \textcolor{red}{-26.54} & \textcolor{red}{-15.37}\\
MoE-Adapter4CL~\cite{moe4cl2024} 
& Architecture
& 52.94 \textcolor{gray}{\scriptsize$\pm$ 3.14}
& 77.48 \textcolor{gray}{\scriptsize$\pm$ 0.91}
& 76.78 \textcolor{gray}{\scriptsize$\pm$ 6.17}
& 87.40 \textcolor{gray}{\scriptsize$\pm$ 2.74}
& \textcolor{blue}{+23.84} & \textcolor{blue}{+9.92} \\
RAPF~\cite{rapf2024} 
& Representation
& 87.88 \textcolor{gray}{\scriptsize$\pm$ 0.12} 
& 92.27 \textcolor{gray}{\scriptsize$\pm$ 0.12}
& 80.40 \textcolor{gray}{\scriptsize$\pm$ 1.57}
& 87.57 \textcolor{gray}{\scriptsize$\pm$ 1.75}
& \textcolor{darkgreen}{-7.48} & \textcolor{darkgreen}{-4.70}\\
SD-LoRA~\cite{sdlora2025} 
& Architecture
& 69.03 \textcolor{gray}{\scriptsize$\pm$ 1.33}
& 87.78 \textcolor{gray}{\scriptsize$\pm$ 0.71}
& 73.39 \textcolor{gray}{\scriptsize$\pm$ 9.60}
& 83.64 \textcolor{gray}{\scriptsize$\pm$ 4.90}
& \textcolor{darkgreen}{+4.36} & \textcolor{darkgreen}{-4.14} \\
\bottomrule
\end{tabular}
}
\end{table*}

\subsubsection{Analysis of PTM-based methods}

The experimental results of PTM-based methods reveals a far more different phenomenon, challenging the conventional wisdom that more memory leads to better results. 
First, across all four datasets, a high-efficiency “sweet spot” emerges in the low-memory range (typically under 20 MB). In this region, methods like CodaPrompt, RanPAC, and InfLoRA achieve state-of-the-art or highly competitive performance at a minimal memory cost. On the challenging CIFAR-100 benchmark, for instance, RanPAC achieves a remarkable 90.59\% accuracy with only 16.0 MB, while CodaPrompt attains 83.78\% with 15.7 MB. This proves that with intelligent mechanisms, exceptional performance is attainable without significant memory overhead.

More significantly, our analysis uncovers a core finding: more memory does not guarantee better performance and can be dramatically inefficient. This is best exemplified by comparing L2P with prompt-based methods. On every benchmark, L2P consumes over 440 MB of memory, yet its performance is consistently outmatched by CodaPrompt, which uses less than 4\% of that memory. Similarly, RanPAC's accuracy on TinyImageNet only improves from 88.33\% to 89.55\% as its memory cost explodes from 16.0 MB to 439 MB, highlighting a severe decline in efficiency. These results strongly suggest that for PTM-based continual learning, the \textit{quality} and \textit{structure} of the stored knowledge are far more critical than the sheer \textit{quantity}. Efficient strategies that learn to query and adapt the vast knowledge already embedded within PTMs (\textit{e.g.}, CodaPrompt's dynamic prompting) are more effective and scalable than those that simply allocate more memory for new parameters.

In summary, our unified memory analysis provides a crucial perspective for the continual learning community. It demonstrates that progress should be measured not just by peak accuracy, but by performance efficiency (accuracy per megabyte). For future research, the focus should shift away from memory-intensive replay or naive parameter expansion, and towards developing sophisticated, low-cost mechanisms to intelligently manage and access knowledge in powerful pre-trained models. This is the key to building truly practical and scalable lifelong learning systems.

\subsection{Investigation 3: Robustness to Semantic Structure in Cross-domain and Category-randomized Settings}\label{subsec:crossdomain_experiment}

To assess algorithmic robustness against varying semantic structures, we conduct experiments on the 5-dataset benchmark~\cite{Ebrahimi2020}\cite{gpm2021}\cite{trgp2022}\cite{l2p2022}\cite{dualprompt2022}\cite{api2023}. We compare performance in two configurations introduced in Section~\ref{subsubsec:Semantic Structure}: the standard cross-domain setting and our proposed category-randomized setting. The latter deliberately breaks this coherence by creating tasks from a shuffled pool of all classes across all domains, directly testing whether models rely on task-level semantic shortcuts. The results in Table~\ref{tab:cross_domain_results} highlight a significant divergence in robustness across the two settings, revealing important characteristics of different algorithmic strategies.

\subsubsection{Analysis of the cross-domain setting}
In cross-domain setting, a clear performance difference emerges between different kinds of methods (Table~\ref{tab:cross_domain_results}). Methods leveraging either explicit data rehearsal or powerful pre-trained models prove most effective at adapting to drastic domain shifts.

Replay-based methods (\textit{e.g.}, ERACE, 85.85\% last acc.) and PTM-based approaches (\textit{e.g.}, RAPF, 87.88\%; RanPAC, 87.10\%) are the top performers. Their success stems from rehearsing past data and adapting pretrained features, respectively. In contrast, regularization-based methods like EWC (26.24\%) and LwF (39.79\%) fail, as their parameter- or function-space constraints are insufficient to bridge the large statistical gaps between domains.

Optimization-based methods like GPM and TRGP yield intermediate results. GPM, for example, achieves a last accuracy of 69.87\%, suggesting that gradient projection can prevent some interference but may overly restrict the model's plasticity, hindering its ability to fully adapt to a new domain.

Finally, the performance of PEFT methods is nuanced. Simpler prompt-based methods like DualPrompt and CodaPrompt deliver strong but not state-of-the-art results. However, more sophisticated PEFT techniques like InfLoRA and RAPF are among the top performers. This demonstrates that the specific adaptation mechanism is more critical for cross-domain success than merely using a PTM.

\subsubsection{Analysis of the category-randomized setting}
\begin{figure}[t]
  \centering
   \includegraphics[width=\linewidth]{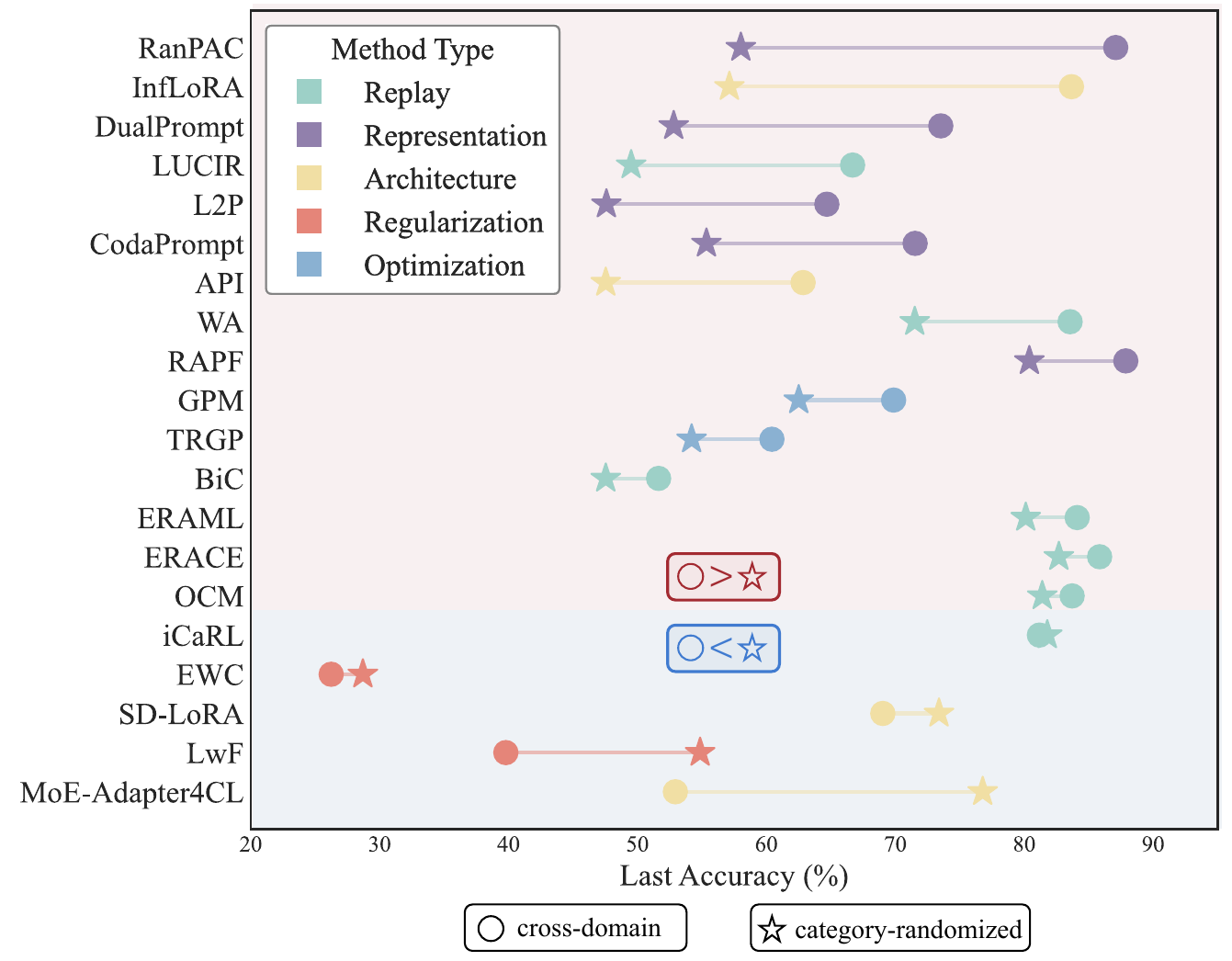}
   \caption{Performance change from the cross-domain (\ding{109}) to the category-randomized (\ding{73}) setting. The plot is divided into a red-shaded region for methods with a performance drop (\ding{109} $>$ \ding{73}) and a blue-shaded region for those with a performance gain (\ding{109} $<$ \ding{73}). Methods are sorted by the magnitude of this change, with the largest drops at the top. Colors denote the algorithmic type (\textit{e.g.}, Replay, Representation).}
   \label{fig:dumbbell}
\vspace{-10pt}
\end{figure}

The category-randomized setting is designed to test genuine robustness by removing the crutch of intra-task semantic coherence. Comparing performance against the cross-domain setting (Table~\ref{tab:cross_domain_results}) reveals three distinct behavioral patterns, particularly when analyzing the change in \textit{Last Accuracy}. We visualize this performance shift in Figure~\ref{fig:dumbbell} to provide a more intuitive understanding of how each method is affected.


\textit{i. Methods exhibiting severe performance degradation.} 
A key observation is the sharp performance decline of several high-performing methods when intra-task semantic coherence is removed. Notably, RanPAC’s accuracy drops by 29.08 percentage points, and similar drops are observed across prompt-based methods (DualPrompt: -20.72\%; L2P: -17.10\%; CodaPrompt: -16.19\%).

This collapse in final task performance suggests these methods heavily exploit task-level semantic regularities as an implicit inductive bias. When tasks contain semantically coherent classes, these methods can learn compact, task-specific representations that capture shared features. However, when forced to simultaneously learn disparate concepts within a single task, their task-level adaptation mechanisms become counterproductive, attempting to find non-existent commonalities among fundamentally unrelated classes. For example, for RanPAC, its single, fixed random projection layer struggles to generate sufficiently distinct feature representations for semantically diverse classes within the same task; for L2P and DualPrompt, their task-level prompts are forced to find a compromised solution for heterogeneous tasks, resulting in suboptimal representations for the unrelated classes.

\textit{ii. Methods demonstrating unexpected improvement.} Counter-intuitively, some methods improved under the more challenging category-randomized conditions. Regularization-based methods like EWC (+3.9\%) and LwF (+15.07\%) have modest gains. This suggests their constraints become more beneficial when task-level semantic coherence is removed, as they prevent the model from over-specializing to spurious task-level patterns. However, their low absolute performance indicates regularization alone is insufficient for such heterogeneous tasks.

The most dramatic improvement is observed in MoE-Adapter4CL, which exhibits an extraordinary 23.84 percentage point increase. This remarkable enhancement reveals a fundamental architectural advantage: the mixture-of-experts framework, originally designed to handle inter-task diversity, inadvertently excels when confronted with intra-task heterogeneity. In the category-randomized setting, each task contains diverse classes from different domains, effectively creating multiple implicit sub-tasks within a single task. The routing mechanism can leverage this diversity by assigning different experts to handle distinct semantic clusters, transforming what appears to be a challenge into an opportunity for specialization. This explains both its under-performance in the homogeneous cross-domain setting and its success here.


\textit{iii. Methods maintaining relative stability.} Several methods demonstrate robustness, with their performance being largely agnostic to the semantic task composition. iCaRL remains remarkably stable, with its last accuracy changing by only +0.61\%. Replay-based methods like ERACE (-3.16\%) and ERAML (-3.99\%) also show high resilience. Other methods, including GPM (-7.36\%) and RAPF (-7.48\%), show moderate but manageable degradation. This stability suggests their mechanisms operate at a level of abstraction (\textit{e.g.}, exemplar replay, gradient projection, or decoupled parameter updates) that is less dependent on the semantic composition of tasks.

Beyond absolute accuracy drops, the category-randomized setting reveals algorithmic fragility in two other ways. First, we observe a significant divergence between Last Accuracy and Average Accuracy for several methods. For example, replay methods like OCM see their Average Accuracy improve (\textit{e.g.}, +9.28\%) while Last Accuracy slightly falls, masking a critical final-task decay. We argue that Last Accuracy remains the more meaningful indicator for real-world capability, as it assesses the model's performance on the entire accumulated knowledge base. 
Second, the standard deviation, calculated over multiple runs with different random seeds (as reported in Table~\ref{tab:cross_domain_results}), increases for several methods. For instance, the variance for BiC (10.34), SD-LoRA (9.60), and InfLoRA (9.25) is notably higher in the category-randomized setting. This increased variance suggests that their performance might be more sensitive to factors like initialization or data ordering when the task structure is less predictable, highlighting a potential area for improving their robustness.



In summary, the category-randomized setting serves as a valuable diagnostic. The divergent performance patterns suggest that a method's success can be closely tied to the semantic structure of the tasks. Strategies that appear highly effective on traditional, semantically coherent benchmarks may not generalize to scenarios where data arrives in a more chaotic, unstructured manner. This underscores the importance of developing methods that are not only accurate but also robust to variations in the underlying semantic organization of the learning curriculum.
\section{Conclusion}\label{sec:conclusion}

In this paper, we present \textit{LibContinual, a unified and reproducible library for continual learning} that re-implements and evaluates the major methods under consistent protocols. 
Our investigations show that training-from-scratch methods collapse under the online learning setting, whereas parameter-efficient adaptations of pre-trained models achieve strong accuracy with modest memory usage. 
We further highlight the necessity of resource-aware evaluation, demonstrating that storage forms and budgets are critical to performance. 
Moreover, the category-randomized setting reveals that many approaches rely heavily on semantic coherence, underscoring the importance of robust strategies for knowledge management in realistic environments. 
By consolidating benchmarks, protocols, and metrics, LibContinual offers a reliable foundation for future research and encourages the development of continual learning methods that balance efficiency, robustness, and applicability.

\section*{Acknowledgments}
This work is supported in part by the National Natural Science Foundation of China (62576160, 62192783), the Young Elite Scientists Sponsorship Program by CAST (2023QNRC001), and the Australian Research Council’s Discovery Project (DP220101784).




\bibliographystyle{IEEEtran}
\bibliography{reference.bib}

\clearpage
\appendices

\section{Detailed Version For A Classic Taxonomy of Continual Learning Methods}
This appendix provides an expanded discussion of the classic taxonomy of continual learning methods introduced in Section III-A of the main paper. It offers a more in-depth review of the core principles, representative works, and a critical analysis for each of the five major categories, thereby offering a richer context for the methodologies evaluated within LibContinual.

To address the challenges posed by continual learning, a wide variety of solutions have emerged from different perspectives in the research community. 
In this section, we begin by systematically reviewing and categorizing these approaches. Drawing upon the taxonomy proposed in a recent comprehensive survey by~\cite{Wang2024}, we broadly classify mainstream continual learning methods into five major categories: regularization-based, replay-based, optimization-based, representation-based, and architecture-based methods.
In the subsequent subsections, we will briefly introduce the core principles and representative works of these categories.
A discussion will be provided for each category, focusing on their distinctive advantages and unresolved limitations.


\subsection{Regularization-based methods}
Regularization-based methods augment the training objective with a penalty term to mitigate forgetting. When learning task $\mathcal{T}_t$, the parameters $\theta_t$ are found by optimizing,
\begin{equation}\small
    \theta_t = \arg\min_{\theta} \mathbb{E}_{(x,y)\sim\mathcal{D}_t}[\mathcal{L}(f_{\theta}(x),y) + \Omega(\theta,\theta_{t-1})].
\end{equation}
The objective balances plasticity, driven by the empirical risk on the new data $\mathcal{D}_t$, with stability, enforced by the regularizer $\Omega(\theta, \theta_{t-1})$ which penalizes deviations from the previous state. The innovation within this family of methods lies in the specific design of the regularizer $\Omega$.

The regularizer $\Omega$ can be defined in the parameter space, focusing on the importance of individual weights~\cite{ewc2017},\cite{MAS2017},\cite{SI2017},\cite{RWALK2018}. For instance, \textit{Elastic Weight Consolidation} (EWC)~\cite{ewc2017} uses a quadratic penalty,
\begin{equation}\small
\Omega(\theta,\theta_{t-1})=\sum_i F_i(\theta_i-\theta_{t-1,i})^2,
\end{equation}
where $\theta_i$ is the value of the $i$-th parameter of the model, and $F_i$ is the diagonal element of the Fisher Information Matrix (FIM). The FIM serves as a proxy for parameter importance by penalizing changes to parameters with a high $F_i$ value.
EWC protects weights critical for past tasks.

Alternatively, $\Omega$ can be defined in the functional space, focusing on preserving the model's input-output behavior~\cite{lwf2016},\cite{Iscen2020},\cite{Castro2018},\cite{Triki2017}. \textit{Learning without Forgetting} (LwF)\cite{lwf2016} employs knowledge distillation on new data to preserve the previous model's outputs. It enforces functional consistency by treating the old model's predictions on new data as soft targets for the new model, defined as,
\begin{equation}\small
    \Omega(\theta,\theta_{t-1})=\mathbb{E}_{x\sim\mathcal{D}_t}[\mathcal{L}_{KD}(f_{\theta_{t-1}}(x),f_{\theta}(x))].
\end{equation}

The core idea of penalizing important changes continues to evolve. For instance, recent work has explored regularization in the spectral domain~\cite{Lewandowski2025} and the topological domain~\cite{Fan2024}. Furthermore, researchers have also begun to adapt and apply these regularization methods to the continual instruction tuning of large multimodal models~\cite{he2023, Zheng2023zscl}.

\textit{Discussion:} Regularization-based methods are foundational to continual learning due to their simplicity and effectiveness. However, in the current landscape dominated by Pre-trained Model (PTM), directly applying a traditional constraint like EWC across the entire network is often suboptimal~\cite{he2023}. PTMs possess powerful, general-purpose representations, and imposing strong global constraints can paradoxically limit their adaptability to new tasks. 
We believe the future value of the regularization approach lies in its flexibility as a strategic component rather than a standalone, global solution. For instance, regularization can be applied more precisely to lightweight modules in the PTM (such as Adapters), preserving the core knowledge of the PTM while enabling targeted adaptation.

\subsection{Replay-based methods}
Replay-based methods address catastrophic forgetting by storing a small subset of past data in a memory buffer and rehearsing it alongside new task data. This strategy directly counteracts the challenges posed by the non-stationary data stream by providing the model with explicit reminders of previously learned knowledge, thereby approximating training on the union of all data seen so far.

When learning the $t$-th task $\mathcal{T}_t$, the optimization objective for replay-based methods is to minimize a composite loss,
\begin{equation}\small
\begin{split}
    \min_{\theta}\mathbb{E}_{(x,y)\sim\mathcal{D}_t}[\mathcal{L}(f_{\theta}(x),y)]&+\beta\cdot\mathbb{E}_{(x_m,y_m)\sim\mathcal{M}}[\mathcal{L}_{m}(f_{\theta}(x_m),y_m)], \\ 
    &\mathcal{M}=\bigcup_{k=1}^{t-1}\mathcal{S}_k. \label{formula:replay}
\end{split}
\end{equation}
In this formulation, $\mathcal{M}$ represents a memory buffer containing a limited set of exemplars $\mathcal{S}_k$ from past tasks. The content of this buffer $\mathcal{M}$, fundamentally defines the specific replay strategy. It may contain raw past examples~\cite{Oord2017}\cite{aljundi2019}\cite{bang2021}\cite{liu2020}, pseudo-data synthesized by a generative model~\cite{shin2017}\cite{ostapenko2019}\cite{cong2020}\cite{van2020}, or abstract latent representations~\cite{hayes2020}\cite{zhu2022}.
Regardless of the strategy, the model is regularized by minimizing a replay loss $\mathcal{L}_m$ on samples drawn from this buffer. The hyperparameter $\beta$ controls the trade-off between learning the new task and preserving old knowledge. The central challenge for these methods lies in how to construct, manage, and utilize the memory buffer 
$\mathcal{M}$ to best approximate the true data distribution of past tasks under strict memory constraints.

A seminal work in this area is iCaRL~\cite{icarl2017}, which populates its memory buffer $\mathcal{M}$ with exemplars selected via a herding process and uses knowledge distillation as the replay loss ($\mathcal{L}_m$ in Eq.~\ref{formula:replay}) to preserve the previous model's outputs. However, a critical issue with simple replay is the severe data imbalance between the large number of new task samples and the few replayed exemplars. This induces a strong predictive bias towards new classes. Subsequent research has focused on mitigating this imbalance~\cite{lucir2019},\cite{wa2020},\cite{bic2019}. For instance, methods like LUCIR~\cite{lucir2019} and BiC~\cite{bic2019} introduce various rebalancing techniques, such as cosine normalization or post-hoc bias correction.

These challenges of representation stability and data imbalance are further amplified in the demanding online CL setting. Here, while simple Experience Replay (ER) serves as a strong baseline, more advanced techniques have emerged. For example, OCM\cite{ocm2022} learns more holistic representations via mutual information maximization to improve feature robustness, while ER-ACE\cite{erace2022} uses an asymmetric loss to prevent the abrupt representation drift common at task boundaries.

A significant recent development is the resurgence of generative replay, a strategy that synthesizes pseudo-data for rehearsal~\cite{shin2017}\cite{Maracani2021}. This trend is catalyzed by powerful diffusion models, which offer a compelling solution to storage and privacy constraints~\cite{Meng2024}\cite{rong2025}. Furthermore, innovations are emerging that replay more abstract forms of knowledge. For instance, Saliency-driven Experience Replay (SER) proposes using a forgetting-free saliency prediction network to modulate and stabilize the features of the main classification model~\cite{Bellitto2024}. Concurrently, a deeper theoretical understanding of replay is emerging, revealing that naive replay can sometimes be detrimental and motivating the design of more intelligent, non-random sampling strategies~\cite{wan2025},\cite{Mahaviyeh2025}.

\textit{Discussion:} Replay-based methods remain at the forefront of continual learning research. Their primary strength lies in the direct, data-driven constraint against forgetting, an intuitive strategy that has proven to be highly effective empirically. While effective, their performance is intrinsically tied to the size and quality of the memory buffer, creating a trade-off between memory overhead, potential privacy risks,  and learning efficacy. We argue that future progress hinges on two key areas: developing more intelligent sampling strategies to maximize the utility of a limited memory budget, and establishing a stronger theoretical foundation to explain why and when replay is most effective. Answering these questions is essential for unlocking the full potential of replay and building truly scalable and robust lifelong learning systems.

\subsection{Optimization-based methods}

Optimization-based methods fundamentally seek to address the stability-plasticity dilemma in continual learning by formulating the learning process as a constrained optimization problem. The core idea is to ensure that updates performed for new tasks do not adversely affect the performance on previously learned tasks~\cite{ogd2020}~\cite{Recursive2022}. 
This principle is elegantly captured by the following constrained optimization objective:
\begin{equation}\small
\begin{aligned}\small
& \theta_t = \arg\min_{\theta} \left[ \mathbb{E}_{(x,y)\sim\mathcal{D}_t}\mathcal{L}(f_{\theta}(x),y)  \right]. 
& \text{s.t.} \langle \nabla_\theta \mathcal{L}_{t}, \nabla_\theta \mathcal{L}_{t-1} \rangle \geq 0 ,
\end{aligned}
\end{equation}
where $\mathcal{L}_{{t}}$ is the loss on the current task, and the constraint enforces that the gradient update for the new task does not increase the loss on previously seen tasks~\cite{owm2019}. 
This formulation encapsulates the essential motivation behind optimization-based continual learning: to find an update direction that is beneficial, or at least non-destructive, to prior knowledge while accommodating new information. 

A foundational approach is GPM~\cite{gpm2021}, which projects gradients of new tasks orthogonally to subspaces spanned by past task features
\begin{equation}\small
\nabla_{\mathbf{W}^{l}} \mathcal{L}_{t} = \nabla_{\mathbf{W}^{l}} \mathcal{L}_{t} - \mathbf{M}^{l} (\mathbf{M}^{l})^{\top} \nabla_{\mathbf{W}^{l}} \mathcal{L}_{t},
\label{eq:gpm}
\end{equation}
where \(\mathbf{M}^{l}\) is the basis matrix of the core subspace of the \(l\)-th layer of the model from past tasks.
While GPM prevents interference, its strict orthogonality limits {plasticity} by restricting new task adaptation.
Adam-NSCL~\cite{comatrix2021} proposes an alternative null space projection strategy grounded in singular value decomposition (SVD). For layer \(l\), it constructs the uncentered feature covariance matrix \(\bar{\mathcal{X}}^{l}_{t-1} = \frac{1}{\tilde{n}_{t-1}}(\bar{X}^{l}_{t-1})^{\top}\bar{X}^{l}_{t-1}\) from previous tasks, where \(\bar{X}^{l}_{t-1}\) concatenates input features of all seen tasks. Through SVD decomposition as below,
\begin{equation}\small
    U^{l},\Lambda^{l},(U^{l})^{\top} = \text{SVD}(\bar{\mathcal{X}}^{l}_{t-1}),
\end{equation}
the method isolates the approximate null space via the singular vector submatrix \(U^{l}_{2}\) corresponding to smallest singular values (\(\lambda \leq a\lambda_{\min}^{l}\)). The gradient projection is then formulated as,
\begin{equation}\small
\nabla_{\mathbf{W}^{l}} \mathcal{L}_{t} = U^{l}_{2} (U^{l}_{2})^{\top} \nabla_{\mathbf{W}^{l}} \mathcal{L}_{t}.
\label{eq:adamnsc}
\end{equation}
This operation forces the gradient update into the null space of \(\bar{\mathcal{X}}^{l}_{t-1}\), ensuring parameter updates satisfy \(\bar{\mathcal{X}}^{l}_{t-1}\Delta w^{l}_{t,s}=0\) and thus \textit{strictly avoid interference} with previous feature representations. Compared to GPM's explicit orthogonality constraint, Adam-NSCL's covariance null space projection provides a more geometrically interpretable solution to forgetting prevention. However, both methods face plasticity limitations due to the constrained update space.

To enhance plasticity, TRGP~\cite{trgp2022} introduces adaptive trust regions. 
For the \(t\)-th task and past tasks \(j\), when the gradient similarity reaches a certain level, the gradient can be projected onto the trust region to improve the learning capability for the new task. The trust region is defined as follows: 
\begin{equation}\small
\mathcal{TR}_{t}^{l} = \left\{ j < t : \frac{ \| \operatorname{Proj}_{\mathcal{S}_{j}^{l}} (\nabla_{\mathbf{W}^{l}} \mathcal{L}_{t}) \|_{2} }{ \| \nabla_{\mathbf{W}^{l}} \mathcal{L}_{t} \|_{2} } \geq \epsilon^{l} \right\}.
\label{eq:tr_def}
\end{equation}
Here, \(\mathcal{S}_{j}^{l}\) is the subspace for task \(j\) at layer \(l\), and \(\epsilon^{l}\) is a similarity threshold. 
TRGP modulates updates using learnable scaling matrices \(\mathbf{Q}_{j,t}^{l}\),
\begin{equation}\small
\begin{aligned}
& \min_{\{ \mathbf{W}^{l} \}_{l}, \{ \mathbf{Q}_{j,t}^{l} \}_{l, j \in \mathcal{TR}_{t}^{l}}}  \mathcal{L} \left( \{ \mathbf{W}_{\text{eff}}^{l} \}_{l}, \mathcal{D}_{t} \right) \\
& \text{s.t.} \quad \mathbf{W}_{\text{eff}}^{l} = \mathbf{W}^{l} + \sum_{j \in \mathcal{TR}_{t}^{l}} \left[ \operatorname{Proj}_{\mathcal{S}_{j}^{l}, \mathbf{Q}} (\mathbf{W}^{l}) - \operatorname{Proj}_{\mathcal{S}_{j}^{l}} (\mathbf{W}^{l}) \right],
\end{aligned}
\label{eq:trgp}
\end{equation}
where \(\operatorname{Proj}_{\mathcal{S}_{j}^{l}, \mathbf{Q}}\) denotes scaled projection operator. 
By performing gradient protection within the trust region and simultaneously utilizing the scaled-projected weights for updates, this method achieves a balance between the model's stability and plasticity.  
Recent work revisits the optimization landscape itself. AdaBOP~\cite{adbop2025} derives a closed-form projection matrix,
\begin{equation}\small
\mathbf{P}_{t-1}^{l} = \left( \mathbf{I} + \lambda \bar{\mathbf{X}}_{t-1}^{l} (\bar{\mathbf{X}}_{t-1}^{l})^{\top} \right)^{-1},
\label{eq:adabop}
\end{equation}
where \(\bar{\mathbf{X}}_{t-1}^{l}\) contains past task features. 
By explicitly constraining plasticity and stability, this method obtains an explicit form of the optimal solution; furthermore, it achieves favorable performance by tuning the hyperparameter \(\lambda\) for each task and each layer.   

\textit{Discussion:} Optimization-based methods are increasingly adopted as plug-and-play modules to mitigate catastrophic forgetting in continual learning systems. However, this trend has overshadowed fundamental innovation in their core anti-forgetting mechanisms—most current implementations rely on conventional gradient constraints without revisiting the underlying optimization principles ~\cite{prompt_nscl2024}. 
We urge renewed focus on re-examining optimization-based approaches themselves, developing novel theories that intrinsically encode forgetting resistance rather than merely applying constraints. Concurrently, emerging memory-efficient implementations like \textit{Adaptive Plasticity Improvement} (API)~\cite{api2023} demonstrate the value of optimizing storage overhead, providing an inspiring direction for practical deployment where resource constraints demand lightweight solutions.

\subsection{Representation-based methods}
Representation-based methods shift the focus of continual learning from ``how to preserve old knowledge" to ``how to learn more essential and universal knowledge". The core philosophy of this approach can be elucidated through a two-phase process. The overall model, denoted as $f_\theta$ in Section 2 in the main text, is decomposed into a feature encoder $f_\text{enc}$ with parameters $\theta_\text{enc}$ and a classifier $f_\text{cls}$ with parameters $\theta_\text{cls}$.

\noindent \textbf{Phase 1} (Representation Learning).
\begin{equation}\small
    \min_{\theta_\text{enc}} \mathcal{L}_{\text{pretext}}(\theta_\text{enc}; \mathcal{D}_{\text{pretrain}}).
\end{equation}

\noindent \textbf{Phase 2} (Continual Learning).
\begin{equation}\small
    \begin{split}
        \min_{\theta_\text{cls}, [\theta_\text{enc}]} & \mathbb{E}_{(x,y) \sim D_t} [\mathcal{L}(f_\text{cls}(f_\text{enc}(x; \theta_\text{enc}); \theta_\text{cls}), y)] \\
        & \text{s.t.} \quad f_\text{enc} \in \mathcal{F}_\text{stable}.
    \end{split}
\end{equation}
In the first phase, a powerful encoder $f_\text{enc}$ is trained on a large-scale dataset $\mathcal{D}_\text{pretrain}$ via a pretext task, optimizing a pretext loss $\mathcal{L}_\text{pretext}$. This phase, which typically involves self-supervised learning or large-scale supervised pre-training~\cite{clip2021,vit2021}, aims to yield a high-quality, universal feature encoder with parameters $\theta_\text{enc}$. While the pre-training in Phase 1 is foundational, most contemporary methods concentrate on the strategies for Phase 2, where the model learns the current task $\mathcal{T}_t$ by minimizing the loss $\mathcal{L}$ on $\mathcal{D}_t$. To prevent catastrophic forgetting, the optimization is constrained such that the encoder $f_\text{enc}$ remains within a stable function space, $\mathcal{F}_\text{stable}$. This special handling of the encoder's parameters, denoted by $\theta_\text{enc}$, leads to two primary strategies.

The most direct and robust strategy to enforce stability is to keep the powerful encoder entirely frozen after Phase 1, \textit{i.e.}, $\theta_\text{enc}$ are fixed. In this case, learning is confined to lightweight modules that operate on these fixed representations.

A leading frozen-encoder paradigm is prompt-based learning, where small, parameter-efficient ``prompts" are learned to instruct the model. The seminal work, \textit{Learning to Prompt} (L2P)~\cite{l2p2022}, introduced a prompt pool and a key-value based query mechanism to select a subset of prompts for each input, effectively storing task-specific knowledge outside the core model. This concept was advanced by DualPrompt~\cite{dualprompt2022} decomposes prompts into ``General" and ``Expert" types to better manage task-invariant and task-specific knowledge inspired by Complementary Learning Systems theory. More recently, CodaPrompt~\cite{codaprompt2023} proposed an attention mechanism over a set of prompt components, enabling the creation of dynamically composed prompts and, critically, facilitating a fully end-to-end optimization of the query-prompt system. An alternative and highly effective approach that also employs a frozen encoder is RanPAC~\cite{ranpac2023}. This method introduces a training-free adaptation mechanism by inserting a frozen, non-linear Random Projection layer after the feature extractor. This layer projects features into a higher-dimensional space to improve their linear separability for a subsequent prototype-based classifier.

To allow for greater plasticity, a second strategy involves cautiously fine-tuning the encoder parameters $\theta_\text{enc}$ while learning new tasks. This re-introduces the risk of forgetting, necessitating methods that carefully balance adaptation with the preservation of the encoder's powerful representations. For instance, SLCA~\cite{Zhang2023} fine-tunes the backbone with a very low learning rate and uses a classifier alignment technique to handle prediction biases. Other methods, like Co2L~\cite{Cha2021}, pair fine-tuning with a self-supervised contrastive distillation loss to explicitly maintain representation stability. More recently, RAPF~\cite{rapf2024} leverages the textual features from a vision-language model (CLIP) to adaptively adjust representations for semantically similar classes, followed by a decomposed parameter fusion strategy on a linear adapter to further mitigate forgetting during the fine-tuning process.

\textit{Discussion:} Representation-based methods, especially those built upon the foundation of large-scale PTMs, currently represent one of promising frontiers of rehearsal-free continual learning. Their strength lies in leveraging flexible feature representation modules to further guide or process pre-trained features, thereby achieving better feature representations. This highlights a key insight: for many continual learning problems, the core challenge is not learning new features from scratch, but rather learning to effectively access and combine the rich features already present in PTMs. Therefore, the central challenges lie in learning more potent feature representations and developing more fine-grained mechanisms to identify and preserve the features essential for preventing forgetting. Furthermore, as the field increasingly moves towards multimodal foundation models, addressing the representation gaps and discrepancies between different modalities, and how to continually learn on them without catastrophic interference, emerges as a vital new frontier.


\subsection{Architecture-based methods}
Architecture-based methods tackle catastrophic forgetting by structurally isolating task-specific knowledge, thereby preventing destructive interference by design. These approaches modify the model's architecture, typically by composing a stable, shared component with expandable, task-specific modules. When learning a new task $\mathcal{T}_t$, the model's parameters are formed by combining a shared backbone $\theta$ with a set of task-specific parameters $\{\hat{\theta}_k\}_{k=1}^t$. The optimization problem for task $\mathcal{T}_t$ is formulated to update only the newly introduced parameters $\hat{\theta}_t$.
\begin{equation}\small
    \begin{split}
        \min_{\hat{\theta}_t} \mathbb{E}_{(x,y) \sim \mathcal{D}_t} [\mathcal{L}(f(x; & \theta \oplus \hat{\theta}_t), y)] + \lambda \cdot \mathcal{R}(\hat{\theta}_t)\\
        & \text{s.t.} \quad \mathcal{R}(\cdot)=||\hat{\theta}_t||_{\text{sparse}}.
    \end{split}
\end{equation}
Here, $\theta$ represents the parameters of the core model, which are typically frozen to preserve generalized knowledge and ensure stability. The term $\hat{\theta}_t$ denotes the set of parameters exclusively allocated for learning task \(T_t\), providing plasticity. The composition operator $\oplus$ signifies how these parameter sets are combined, such as through masking, additive decomposition, or modular routing. Finally, the regularization term $\mathcal{R}(\cdot)$ is often used to enforce constraints like sparsity on $\hat{\theta}_t$, ensuring the model's growth is scalable and parameter-efficient.

Early implementations of this principle focused on parameter isolation within a fixed-capacity model~\cite{Yoon2018}\cite{Serra2018}\cite{Mallya2018}\cite{Oswald2020}\cite{Aljundi2017}. Methods like \textit{Hard Attention to the Task} (HAT)~\cite{Serra2018} and PackNet~\cite{Mallya2018} define $\hat{\theta}_t$ as a binary mask applied to $\theta$, effectively creating dedicated sub-networks for each task by freezing important weights from past tasks. Other approaches dynamically expand the architecture. For instance, \textit{Adaptive Plasticity Improvement} (API)~\cite{api2023} evaluates the model's plasticity for a new task and adaptively expands $\hat{\theta}_t$ by adding new neural units if the current plasticity is deemed insufficient. These methods, while effective, were often designed for models trained from scratch.

With the advent of large-scale pre-trained foundation models, $\theta$ is now commonly a powerful, frozen backbone like a Vision Transformer (ViT). Consequently, $\hat{\theta}_t$ is realized through various Parameter-Efficient Fine-Tuning (PEFT) techniques. A prominent strategy is to use \textit{Low-Rank Adaptation} (LoRA). For example, \textit{Interference-Free LoRA} (InfLoRA)~\cite{inflora2024} designs the LoRA matrices (a form of $\hat{\theta}_t$) to lie in a subspace that is orthogonal to the gradients of previous tasks, thereby explicitly minimizing interference. Building on this, \textit{Scalable Decoupled LoRA} (SD-LoRA)~\cite{sdlora2025} decouples the learning of the magnitude and direction of LoRA components. By fixing previously learned directions and only learning new directions alongside all magnitudes, it traces a low-loss path that converges to a shared solution space for all tasks, uniquely enabling rehearsal-free and inference-efficient CL without needing task-specific component selection. Another sophisticated approach, \textit{Mixture-of-Experts Adapters for CL} (MoE-Adapter4CL)~\cite{moe4cl2024}, implements $\hat{\theta}_t$ as a set of experts managed by a task-specific router. This method enhances scalability and leverages a Distribution Discriminative Auto-Selector (DDAS) to automate task identification, preserving the zero-shot capabilities of the underlying vision-language model for out-of-distribution inputs.

\textit{Discussion:} Architecture-based methods, especially those integrated with PEFT on foundation models, currently represent a highly promising frontier in continual learning. Their core strength lies in providing a structural solution to the stability-plasticity dilemma by freezing the general knowledge base ($\theta$) while allowing targeted, efficient updates via $\hat{\theta}_t$. However, a critical challenge emerges from their reliance on task identity. While highly effective in TIL where the task ID is provided, many methods struggle in the more realistic CIL setting. Without an explicit task oracle, dynamically selecting or activating the correct task-specific parameters ($\hat{\theta}_t$) for a given input becomes a non-trivial problem, potentially leading to significant performance degradation. Moreover, as the number of tasks grows, naively accumulating task-specific parameters can lead to a linear or super-linear growth in model size, posing significant memory and computational burdens. The future of this domain lies in developing more sophisticated strategies for managing $\hat{\theta}_t$. Instead of simple expansion, the focus is shifting towards intelligent parameter reuse, composition, and merging.

\section{LibContinual Framework}
This appendix details the software architecture and design principles of the LibContinual framework. It elaborates on the functionality of each core module—from configuration and data handling to algorithm implementation and evaluation, providing a technical blueprint for researchers interested in utilizing or extending the toolbox for their own work.

LibContinual is a comprehensive framework for continual learning, with its overall architecture illustrated in Figure~2 in the main text. To accommodate the integration of various continual learning algorithms within a unified framework, LibContinual is organized into multiple modules. This modular design enables flexible composition and significantly simplifies the development process, making it more manageable and systematic.

\subsection{Config}
The configuration module of LibContinual is implemented using the YAML file format to specify parameters related to data, learning methods, and other experimental settings. A wide range of experimental variables can be defined using key-value pairs, including the continual learning algorithm to be employed, the architecture of the backbone network, dataset paths, among other critical information. To reduce redundant specification of commonly used parameters such as optimizers and backbone architectures, a set of default configuration files is also provided. These default files are first loaded and their contents used as baseline parameters. Subsequently, the custom configuration file is read, and its values are used to update the defaults. The final configuration for the experiment is thus generated through the merging of both default and customized settings.

\subsection{Continual Learner}
The Continual Learner module serves as the core component of LibContinual, responsible for orchestrating the entire continual learning process. Logically, the training procedure can be divided into several key stages along a temporal axis. 

\textbf{Initialization Stage.} This stage handles the setup of all essential components, including logger initialization, configuration file parsing, data loader preparation, algorithm and optimizer instantiation, backbone network construction, and GPU allocation. 

\textbf{Training Stage.} This stage can be further subdivided into pre-task processing, task-specific training, and post-training processing, each allowing for the injection of algorithm-specific logic as needed. 

\textbf{Evaluation Stage.} This stage assesses the model's overall performance on the test set. 

\textbf{Saving Stage.} After the entire training process is completed, this stage is responsible for saving relevant artifacts, such as configuration files, checkpoints, and training logs. 

This logical decomposition enables the flexible insertion of different algorithms into appropriate stages of the pipeline. From a temporal perspective, each task in a continual learning scenario corresponds to a full cycle from pre-task processing to evaluation. The entire experiment proceeds by repeating this cycle for each task following the initial setup. After initialization, the Trainer module executes the training loop to iteratively carry out this process.


 \subsection{Datasets}
Compared to traditional deep learning, where all data are encapsulated in a single data loader, LibContinual introduces a specialized data loader tailored to the unique multi-task setting of continual learning. This data loader is designed to accommodate various requirements specific to continual learning, such as task-wise dataset partitioning and merging. Specifically, LibContinual assumes that all datasets follow a unified directory structure, consisting of two folders named ``train" and ``test", which contain all training and testing images, respectively. Within each of these folders, images belonging to different classes are stored in separate subdirectories. To facilitate rapid and convenient experimentation, LibContinual provides pre-processed dataset archives for several commonly used benchmarks, including CIFAR-10, CIFAR-100, CUB200, ImageNet-R, and Tiny-ImageNet. In the context of continual learning, dataset partitioning is typically defined by two parameters: \textit{init\_cls\_num} and \textit{inc\_cls\_num}, which denote the number of classes in the initial task and in each subsequent incremental task, respectively. In LibContinual, an entire dataset is encapsulated using a ContinualDatasets object. During the initialization phase, this object generates a class order based on the configuration file, and then partitions the dataset into multiple sub-datasets according to the specified \textit{init\_cls\_num} and \textit{inc\_cls\_num}. Furthermore, since evaluation in continual learning often requires testing across multiple task-specific test sets, LibContinual also supports general-purpose operations such as merging and splitting datasets.

\subsection{Backbone}
Backbone networks play a pivotal role in the field of deep learning, and in some cases, the introduction of a new network architecture can significantly advance an entire research area. In LibContinual, a variety of widely adopted backbone models in continual learning are integrated into the Backbone module, including the classical ResNet family, Vision Transformers, and CLIP networks. Moreover, since certain methods may require modifications to the backbone, switching the network structure can be easily achieved by making simple adjustments in the configuration file of LibContinual. A complete model typically consists of two components, a backbone network and a classifier. In most continual learning approaches, the classifier is implemented as a simple linear layer. LibContinual encapsulates these classifiers and provides a set of general-purpose functional interfaces and parameters, thereby reducing redundancy during the development process.

\subsection{CL Algorithm}
For the implementation of specific method modules, several core functionalities are required. Before the training of the current task begins, the \textit{before\_task} function is invoked to perform preliminary operations such as variable initialization, model structure adjustments, and training parameter configuration. During the training phase, the observe function is called with a batch of training samples as input. This function returns the predicted results, classification accuracy, and forward loss. It focuses on how the model processes a batch of data during training, specifically, how the loss is computed and how parameters are updated. In the inference phase, the inference function is invoked with a batch of test samples and returns classification results along with accuracy. This function is concerned with how the model performs forward inference during evaluation. After the training of each task is completed, the \textit{after\_task} function is executed to handle post-task adjustments, such as modifications to the model architecture or memory buffer. This step typically requires user-defined logic tailored to the specific method.

\subsection{Metric}
The module implements commonly used performance metrics in continual learning, such as average task accuracy, backward transfer, forgetting measure, and overall average accuracy. These metrics are used to comprehensively evaluate the model’s performance across different tasks, the effectiveness of knowledge transfer, and the extent of forgetting.
\begin{table*}[htbp]
\centering
\caption{Memory Usage Calculation Examples of Continual Learning Methods.}
\label{tab:comprehensive_analysis_focused}
\resizebox{\textwidth}{!}{%
\begin{tabular}{@{}l c |c c c c c| r c@{}}
\toprule
\textbf{Method} & \textbf{Memory Classification} & \textbf{Image} & \textbf{Feature} & \textbf{Model} & \textbf{Parameter} & \textbf{Prompt} & \textbf{\begin{tabular}[c]{@{}r@{}}Total Memory\end{tabular}} & \textbf{\begin{tabular}[c]{@{}c@{}}Frozen Params\end{tabular}} \\
\midrule

iCaRL & \begin{tabular}[c]{@{}c@{}}Image-based, \\ Model-based\end{tabular} & 32x32x3x1 & - & 472,756 x 4 & 472,756 x 4 & - & \begin{tabular}[c]{@{}r@{}}9,926,048 \\ (9.93 M)\end{tabular} & - \\
\addlinespace

GPM & Feature-based & - & \begin{tabular}[c]{@{}l@{}}$(48^2+576^2+512^2$ \\ $+1024^2+2048^2)$ \\ $\times 4$\end{tabular} & - & 6,704,128 x 4 & - & \begin{tabular}[c]{@{}r@{}}50,172,928 \\ (50.17 M)\end{tabular} & - \\
\addlinespace

L2P & Prompt-based & - & - & - & - & 46,080 x 4 & \begin{tabular}[c]{@{}r@{}}491,920 \\ (0.49 M)\end{tabular} & 491,920 \\
\addlinespace

MoE-Adapter4CL & Parameter-based & - & - & - & 4,104,292 x 4 & - & \begin{tabular}[c]{@{}r@{}}16,417,168 \\ (16.42 M)\end{tabular} & 16,417,168 \\

\bottomrule
\end{tabular}%
}
\end{table*}

\section{Memory Calculation}
This appendix outlines the specific protocol used to implement the Unified Memory Budget, a central component of our investigation into algorithmic efficiency as introduced in Section V-C of the main paper. It details the precise methodology for quantifying the additional memory costs associated with diverse continual learning strategies, thereby establishing the standardized and equitable basis for comparison used in our storage-centric analysis.

To ensure a fair and standardized comparison across continual learning strategies that rely on qualitatively different forms of stored knowledge, we established a unified memory accounting protocol. In our framework, the total memory footprint of a method is calculated as the sum of the static memory required by the network backbone and any additional memory consumed by the specific continual learning algorithm for storing extra information. For the purpose of creating a single, standardized metric, we quantify the entire memory footprint in terms of integer (\texttt{int}) units. The additional storage, which is the dynamic component that varies between methods, is also meticulously quantified using the same base unit. This extra storage encompasses all forms of preserved knowledge as categorized in our storage-centric taxonomy, including the raw pixel values of buffered exemplars (image-based), the elements of stored feature vectors (feature-based), the parameters of saved model snapshots (model-based), the weights of dynamically added network modules (parameter-based), and the tokens of learnable prompts (prompt-based). By converting all these disparate data types into a common integer-based unit of account, this protocol enables an equitable and direct comparison of the true resource efficiency of each method, moving beyond simple performance metrics to reveal the underlying cost-benefit trade-offs.

As shown in Table~\ref{tab:comprehensive_analysis_focused}, our protocol is best illustrated through a concrete application. For instance, calculating the memory footprint of a hybrid method like {iCaRL} involves quantifying and summing each of its storage components. The primary memory cost is from its image-based exemplar buffer (2000 images), with additional allocations for its model-based component (a stored model for knowledge distillation) and the trainable parameters of the active backbone. By converting the storage for each component into our unified integer-based metric, we arrive at its total memory budget of approximately 9.93 M units. This systematic, component-wise summation is applied uniformly across all evaluated methods to ensure a fair and direct comparison of their memory overhead.



 
\vspace{11pt}

\vfill

\end{document}